\documentclass[]{article}
\usepackage{lmodern}
\usepackage[final, nonatbib]{neurips_2023}

\usepackage[numbers]{natbib}
\renewcommand{\bibname}{References}

\usepackage{scalefnt, letltxmacro}
\usepackage[acronym, nowarn]{glossaries}
\glsdisablehyper \makeglossaries
\usepackage{xspace}
\usepackage{float}
\usepackage{physics}
\usepackage[normalem]{ulem}
\usepackage[english]{babel}

\usepackage{enumitem}

\usepackage{amssymb}
\usepackage{mathtools}
\usepackage{amsfonts}
\usepackage{amsmath}
\usepackage{amsthm, thmtools}
\usepackage{booktabs}
\usepackage[]{microtype}
\usepackage{array}
\usepackage[]{lipsum}

\usepackage[linesnumbered, ruled, vlined]{algorithm2e}

\SetCommentSty{mycommfont} \SetKwInput{KwInput}{Input} \SetKwInput{KwOutput}{Output} \SetKwComment{Comment}{/* }{ */}

\usepackage[usenames, dvipsnames, svgnames, table]{xcolor}
\usepackage{xkcdcolors}
\definecolor{pink3}{cmyk}{0, 0.7808, 0.4429, 0.1412}
\usepackage[colorlinks, linktoc=all]{hyperref}
\usepackage[all]{hypcap}
\hypersetup{citecolor=pink3}
\hypersetup{linkcolor=pink3}
\hypersetup{urlcolor=pink3}

\usepackage[capitalize, nameinlink,]{cleveref} %
\crefname{section}{\S}{\S\S} \Crefname{section}{\S}{\S\S}

\definecolor{shadecolor}{gray}{0.9}

\definecolor{mylightgray}{gray}{0.94}

\makeatletter \@ifundefined{c@rownum}{%
\let\c@rownum\rownum }{} \@ifundefined{therownum}{%
\def\therownum{\@arabic\rownum}%
}{} \makeatother

\makeatletter
\newcommand*{\addFileDependency}[1]{%
\typeout{(#1)} \@addtofilelist{#1} \IfFileExists{#1}{}{\typeout{No file #1.}} }
\makeatother

\usepackage[font=footnotesize, labelfont=it, tableposition=top]{caption}
\usepackage{tikz}
\usepackage{graphicx}
\usepackage[]{subcaption} %

\usepackage[export]{adjustbox}

\usepackage{pgfplots}
\pgfplotsset{compat=1.6}
\usepgfplotslibrary{groupplots}
\tikzstyle{every picture}
+=[font=\sffamily]
\tikzstyle{optimized}
= [circle,fill=white,draw=black, dashed,inner sep=1pt, minimum size=20pt, font=\fontsize{10}{10}\selectfont, node distance=1]
\pgfkeys{/pgf/number format/.cd, 1000 sep={}}
\pgfplotsset{
  tick label style={font=\small\sffamily},
  every axis label/.append style={font=\scriptsize\sffamily},
  typeset ticklabels with strut,
}

\pgfplotsset{
  every axis/.append style={ every x tick label/.append style={font=\scriptsize\sffamily, yshift=.5ex,}, every y tick label/.append style={font=\scriptsize\sffamily, xshift=.5ex}, every y label/.append style={xshift=10ex, font=\footnotesize\sffamily}, every x label/.append style={yshift=3ex, font=\footnotesize\sffamily}, every title/.append style={font=\scriptsize\sffamily}, axis lines*=left, },
}
\pgfplotsset{every axis title/.append style={yshift=-1ex, font=\scriptsize\sffamily}}
\pgfplotsset{every axis y label/.append style={yshift=-2ex}}
\pgfplotsset{
  xticklabel={\scriptsize$\mathsf{\pgfmathprintnumber{\tick}}$},
  yticklabel={\scriptsize$\mathsf{\pgfmathprintnumber{\tick}}$},
}
\pgfplotsset{every axis title/.append style={yshift=-1ex}}
\pgfplotsset{/pgfplots/group/.cd, horizontal sep=0.5cm, vertical sep=0.5cm}

\usepgfplotslibrary{external}

\usepackage[
  colorinlistoftodos,
  textsize=scriptsize,
  linecolor=red!30,
  bordercolor=red!30,
  backgroundcolor=red!10
]{todonotes}
\renewcommand{\todo}[2][]{\tikzexternaldisable\@todo[#1]{#2}\tikzexternalenable}

\usepackage[most]{tcolorbox}
{\begin{tcolorbox}[toprule=2mm,left=4pt,right=4pt,top=0pt,bottom=1pt,boxsep=2pt, before skip = 2ex, after skip =2ex]}{\end{tcolorbox}}
\tcbset{boxsep=4pt,left=2pt,right=2pt,top=-0pt,bottom=0pt}

\usepackage{url}

\usepackage{subcaption}

\usepackage{tikz}
\usepackage{graphicx}
\usepackage{amsmath}
\usepackage{amssymb}
\usepackage{bm}
\usepackage{amsthm} %

\declaretheorem[name=Definition]{definition} \declaretheorem[name=Conjecture]{conjecture}

\usepackage{adjustbox}
\usepackage{multirow}
\usepackage{mathtools}

\usepackage{xspace}

\newacronym{SGD}{\textsc{sgd}}{stochastic gradient descent}%
\newacronym{MAP}{\textsc{map}}{maximum-a-posteriori}%
\newacronym{MLE}{\textsc{mle}}{maximum likelihood estimation}%
\newacronym{MNLL}{\textsc{mnll}}{mean negative log-likelihood}%
\newacronym{NLL}{\textsc{nll}}{negative log-likelihood}%
\newacronym{LL}{\textsc{ll}}{log-likelihood}%
\newacronym{RMSE}{\textsc{rmse}}{root mean square error}%
\newacronym{ECE}{\textsc{ece}}{expected calibration error}%
\newacronym{SNR}{\textsc{snr}}{signal-to-noise ratio}%
\newacronym{FID}{\textsc{fid}}{Fr\'echet Inception Distance}%
\newacronym{BPD}{\textsc{bpd}}{bit per dimension}%
\newacronym{NFE}{\textsc{nfe}}{neural function evaluations}%

\newacronym{AE}{\textsc{ae}}{autoencoder}%
\newacronym{WAE}{\textsc{wae}}{Wasserstein Autoencoder}%
\newacronym{VAE}{\textsc{vae}}{Variational Autoencoder}%
\newacronym{BAE}{\textsc{bae}}{Bayesian autoencoder}%
\newacronym{CDF}{\textsc{cdf}}{cumulative density function}%
\newacronym{GAN}{\textsc{gan}}{Generative Adversarial Network}%
\newacronym{DPGMM}{\textsc{dpgmm}}{Dirichlet process Gaussian mixture model}%
\newacronym{MC}{mc}{Monte Carlo}%
\newacronym{SDE}{sde}{Stochastic Differential Equation}%
\newacronym{CNF}{cnf}{Continuous Normaxlizing Flow}%
\newacronym{ODE}{ode}{Ordinary Differential Equation}%
\newacronym{MCMC}{\textsc{mcmc}}{Markov chain Monte Carlo}%
\newacronym{HMC}{\textsc{hmc}}{Hamiltonian Monte Carlo}%
\newacronym{MH}{mh}{Metropolis-Hastings}%
\newacronym{NUTS}{nuts}{no-u-turn sampler}%
\newacronym{SGHMC}{\textsc{sghmc}}{stochastic gradient Hamiltonian Monte Carlo}%
\newacronym[longplural=deep Gaussian processes]{DGP}{\textsc{dgp}}{deep Gaussian process} %
\newacronym{GPLVM}{gplvm}{Gaussian process latent variable model} %
\newacronym{DPMM}{dpmm}{Dirichlet Process Mixture Model} %
\newacronym{VFE}{vfe}{variational free energy} %
\newacronym[longplural=Gaussian Processes]{GP}{\textsc{gp}}{Gaussian Process} %
\newacronym{VI}{\textsc{vi}}{variational inference}%
\newacronym{SVI}{\textsc{svi}}{stochastic variational inference}%
\newacronym{ELBO}{\textsc{elbo}}{evidence lower bound}%
\newacronym{NELBO}{\textsc{nelbo}}{negative evidence lower bound}%
\newacronym{ELL}{\textsc{ell}}{expected log likelihood}%
\newacronym{KL}{\textsc{kl}}{Kullback-Leibler}%
\newacronym{AUC}{auc}{area under the curve}%
\newacronym{BNN}{\textsc{bnn}}{Bayesian neural network}%
\newacronym{DNN}{\textsc{dnn}}{deep neural network}%
\newacronym{CNN}{\textsc{cnn}}{convolutional neural network}%
\newacronym{MLP}{\textsc{mlp}}{multilayer perceptron}%
\newacronym{FRN}{\textsc{frn}}{filter response normalization}%
\newacronym{LN}{\textsc{ln}}{LayerNorm}%
\newacronym{NN}{nn}{neural network}%
\newacronym{RELU}{ReLU}{rectified linear unit}%
\newacronym{NF}{nf}{normalizing flow}%
\newacronym{RBF}{rbf}{radial basis function}%
\newacronym{ARD}{ard}{automatic relevance determination}%
\newacronym{RKHS}{rkhs}{reproducing kernel Hilbert space}%
\newacronym{OT}{ot}{optimal transport}%
\newacronym{WD}{wd}{Wasserstein distance}%
\newacronym{SWD}{swd}{sliced-Wasserstein distance}%
\newacronym{DSWD}{dswd}{distributional sliced-Wasserstein distance}%
\newacronym{LAP}{\textsc{lap}}{linear assignment problem}%
\newacronym{SOLAP}{\textsc{solap}}{sum of bilinear assignment problems}%

\newcommand{\mathbold}[1]{{\boldsymbol{{#1}}}}

\newcommand{\g}{\,|\,}

\newcommand{\nestedmathbold}[1]{{\mathbold{#1}}}

\newcommand{\mbb}{\nestedmathbold{b}}

\newcommand{\mbf}{\nestedmathbold{f}}

\newcommand{\mbm}{\nestedmathbold{m}}

\newcommand{\mbs}{\nestedmathbold{s}}

\newcommand{\mbx}{\nestedmathbold{x}}
\newcommand{\mby}{\nestedmathbold{y}}

\newcommand{\mbA}{\nestedmathbold{A}}
\newcommand{\mbB}{\nestedmathbold{B}}

\newcommand{\mbI}{\nestedmathbold{I}}

\newcommand{\mbM}{\nestedmathbold{M}}

\newcommand{\mbP}{\nestedmathbold{P}}

\newcommand{\mbS}{\nestedmathbold{S}}

\newcommand{\mbW}{\nestedmathbold{W}}
\newcommand{\mbX}{\nestedmathbold{X}}
\newcommand{\mbY}{\nestedmathbold{Y}}

\newcommand{\mbnu}{\nestedmathbold{\nu}}

\newcommand{\mbtheta}{\nestedmathbold{\theta}}

\newcommand{\mbvarepsilon}{\nestedmathbold{\varepsilon}}

\newcommand{\mbzero}{\nestedmathbold{0}}

\newcommand{\Lelbo}{\cL_{\textsc{elbo}}}

\DeclareRobustCommand{\KL}[2]{\ensuremath{\textsc{kl}\left[#1\;\|\;#2\right]}}
\DeclarePairedDelimiterX{\infdivx}[2]{[}{]}{%
#1\;\delimsize\|\;#2%
}

\newcommand{\nmc}{{N_{\mathrm{MC}}}}

\DeclareMathOperator*{\argmax}{arg\,max}
\DeclareMathOperator*{\argmin}{arg\,min}
\DeclareMathOperator*{\logdet}{log\,det}

\newcommand{\eqdef}{\stackrel{{\scriptsize\rm def}}{=}}

\newcommand{\cB}{\mathcal{B}}
\newcommand{\cD}{\mathcal{D}}
\newcommand{\cL}{\mathcal{L}}
\newcommand{\cN}{\mathcal{N}}
\newcommand{\cP}{\mathcal{P}}
\newcommand{\cQ}{\mathcal{Q}}

\newcommand{\cT}{\mathcal{T}}

\newcommand{\cW}{\mathcal{W}}

\newcommand{\E}{\mathbb{E}}

\newcommand{\bbR}{\mathbb{R}}
\newcommand{\bbS}{\mathbb{S}}
\newcommand{\bbE}{\mathbb{E}}

\newcommand{\inv}{{-1}}

\newcommand{\diag}{\textrm{diag}}

\newcommand{\Din}{{D_{\text{in}}}}
\newcommand{\Dout}{{D_{\text{out}}}}

\title{%
On permutation symmetries in Bayesian neural network posteriors: a variational perspective}

\author{Simone Rossi, Ankit Singh, Thomas Hannagan}
\author{%
Simone Rossi \\ Stellantis, France \\
\And Ankit Singh \\ Stellantis, India \\ \And Thomas Hannagan \\ Stellantis, France
}

\begin{document}
  \maketitle

  \begin{abstract}

The elusive nature of gradient-based optimization in neural networks is tied to their loss landscape geometry, which is poorly understood. However recent work has brought solid evidence that there is essentially no loss barrier between the local solutions of gradient descent, once accounting for weight-permutations that leave the network's computation unchanged. This raises questions for approximate inference in Bayesian neural networks (BNNs), where we are interested in marginalizing over multiple points in the loss landscape.
In this work, we first extend the formalism of marginalized loss barrier and solution interpolation to BNNs, before proposing a matching algorithm to search for linearly connected solutions. This is achieved by aligning the distributions of two independent approximate Bayesian solutions with respect to permutation matrices. We build on the results of Ainsworth et al. (2023), reframing the problem as a combinatorial optimization one, using an approximation to the sum of bilinear assignment problem. We then experiment on a variety of architectures and datasets, finding nearly zero marginalized loss barriers for linearly connected solutions. 
  \end{abstract}

  \section{Introduction}
\label{sec:intro}

Throughout the last decade, \glspl{DNN} have achieved significant success in a wide range of practical applications,
becoming the fundamental ingredient for e.g., computer vision \citep[e.g.,][]{Krizhevsky2012,Dosovitskiy2021,He2016,Liu2022},
language models \citep[e.g.,][]{Devlin2019,Radford2018,Brown2020} and generative models \citep[e.g.,][]{Kingma14,Sohl2015,Song2021,Song2020b,Tran2021,Franzese2023}.
Despite recent important advancements, understanding the loss landscape of \glspl{DNN} is still challenging. The characterization
of its highly non-convex nature, its relation with architectural choices like depth and width and the connection with
optimization and generalization are just some of the problems which have been the focus of extensive research in the
last few years \citep[e.g.,][]{Neyshabur2018,Draxler2018a,Garipov2018,Frankle2020a,Ainsworth2023,Entezari2022,Godfrey2022,Pittorino2022}.
It is well known, for example, that one of the fundamental characteristics of deep neural networks is their ability to learn
hierarchical features, and in this regards deeper networks seem to be exponentially more expressive than shallower
models \citep[e.g.,][]{Zeyuan2019,Arora2019,Bai2020,Bengio2005,Chen2020,Yehudai2019}, leading the loss landscape to have
many optima due to symmetries and over-parameterization \citep{Neyshabur2018,Draxler2018a,Zhang2017,Simsek2021a}. At the
same time, the role of the depth of a model in relation with its width is far less understood \citep{Pleiss2021},
despite wide neural networks exhibiting important theoretical properties in their infinite limit behavior \citep[e.g.,][]{Neal1994,Matthews2018,Dutordoir2021,Khan2019,Jacot2018,Garriga-Alonso2018,Cutajar17}.

Two notions that have been useful to shed light on the geometry of loss landscapes are that of \emph{loss barriers} and
\emph{mode connectivity} \citep{Garipov2018,Draxler2018a}.
The mode connectivity hypothesis states that given two points in the landscape, there exists a path connecting them such
that the loss is constant or near constant (or, said differently, the loss barrier is null).
We refer to \emph{linear mode connectivity} when the path connecting the two solutions is linear \citep{Frankle2020a}.
Recently, evidence has surfaced that \gls{SGD} solutions to the loss minimization problem can be linearly connected. Indeed,
\citet{Entezari2022} discuss the role of permutation symmetries from a loss connectivity viewpoint, conjecturing the
possibility that mode connectivity is actually linear once accounting for all permutation invariances. Additionally,
\citep{Ainsworth2023} gathers compelling empirical evidence across several network architectures and tasks, that under
such a permutation symmetry the loss landscape often contains a single, nearly convex basin.

\begin{figure}
  \centering
  \includegraphics[width=\textwidth]{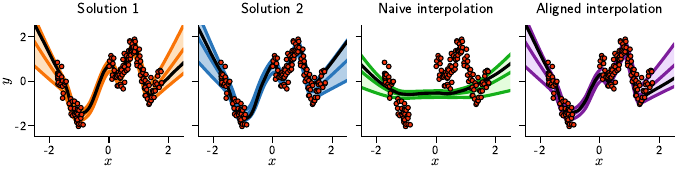}
  \caption{\textbf{Permutation symmetries for regression on the Snelson dataset \citep{Snelson2006}.} \textit{(Left)}
  Two different solutions, with similar function-space behavior (showing $\mu\pm2\sigma$). \textit{(Right)} Functions obtained
  when using two different strategies to interpolate between solutions. When we align the solutions, by taking into account
  the permutation symmetries, we retain the capability to model the data, indicating that there are solutions that once linearly
  interpolated exhibit no loss in performance. Note that, in weight space, the solution found with alignment
  is neither equal to solution 1 nor 2 (see the black curve which is a single function sample with fixed randomness).
  }
  \label{fig:1d_mpl}
\end{figure}
In this work, we are taking a different perspective on this analysis. We are interested in the Bayesian treatment of neural
networks, which results in a natural form of regularization and allows to reason about uncertainty in the predictions
\citep{Tishby1989,Neal1994,MacKay1992}.
Bayesian inference for deep neural networks is notoriously challenging, as we wish to marginalize over multi-modal distributions
with high dimensionality \citep{Izmailov21}. For this reason, there are various ways to approximate the posterior, involving
techniques like variational inference \citep{Graves2011,Blundell2015,Gal2016,Li2017,Osawa2019}, \gls{MCMC} methods
\citep{Neal2011,Metropolis1949,Duane1987}, possibly with stochastic gradients \citep{Cheni2014,ZhangLZCW20,Franzese2021a,Welling2011,Mandt2017}
and the Laplace approximation \citep{MacKay1998,MacKay1998,Ritter2018}.
Indeed, fundamentally the Bayesian posterior and the loss landscapes are tightly interconnected: (i) solutions to the loss
minimization problem are equivalent to \gls{MAP} solutions, (ii) the loss landscape is equivalent to the un-normalized negative
log-posterior. While in theory, given a dataset, the posterior is unique and the solution is global, many approximations
will only explore local properties of the true posterior\footnotemark. \footnotetext{By local properties, we mean that despite
theoretical convergence guarantees of many methods like variational inference and \gls{MCMC}, in practice the true posterior
for deep neural networks is still highly elusive; see for instance the empirical convergence analysis of \gls{HMC} in \citep{Izmailov21}.}
It's worth noting that a posterior over the parameters of the neural network induces a posterior on the functions generated
by the model. %
Permutation symmetries play an important role in the geometry of the weight-space posterior, which are generally not
reflected in function-space. While it is possible to carry out inference directly in function space, this poses a number
of challenges \citep{Sun2019,Matthews2016,Rudner2022,Lee2018}. \cref{fig:1d_mpl} illustrates this situation for a
regression task on the Snelson dataset using a 3-layer \gls{DNN}: on the left we compare two (approximate) solutions
which have different weight-space posterior but similar function-space behavior. Notably, when we interpolate these two
solutions (\cref{fig:1d_mpl} on the right), we completely lose all capability of modeling the data. However, when we account
for permutation symmetries in the posterior, we end up with solutions that once interpolated are still good
approximations. This suggests that for any weight-space distribution, there exists a class of solutions which are
functionally equivalent and linearly connected.
This example motivates an informal generic statement:

\emph{Solutions of approximate Bayesian inference for neural networks are linearly connected after accounting for
functionally equivalent permutations. }

While being similar to the one in \citep{Entezari2022,Ainsworth2023}, if this was to hold true for approximate
\glspl{BNN} it would represent an important step in further characterizing the properties of the Bayesian posterior and
the effect of various approximations. We purposely leave the previous statement broadly open regarding the choice of the
approximation method to allow for a more general discussion. More specifically, in this paper we will analyze and focus our
discussion on the variational inference framework, making a more specific conjecture:
\begin{conjecture}
  \label{claim:lmc}%
  Solutions of \underline{variational inference solutions} for Bayesian inference for neural networks are linearly connected
  after accounting for functionally equivalent permutations.
\end{conjecture}

\begin{minipage}{.6\textwidth}
  \paragraph{Contributions.}
  With this work, we aim at studying the linear connectivity properties of approximate solutions to the Bayesian inference
  problem and we make several contributions.
  (i) We extend the formalism of loss barrier and solution interpolation to \glspl{BNN}. (ii) For the variational
  inference setting, propose a matching algorithm to search for linearly connected solutions by aligning the distributions
  of two independent solutions with respect to permutation matrices. Inspired by \citep{Ainsworth2023}, we frame the
  problem as a combinatorial optimization problem using approximation to the linear sum assignment problem. (iii) We then
  experiment on a variety of architectures and datasets, finding nearly zero-loss barriers for linearly connected
  solutions. In \cref{fig:posterior_cifar} we present a sneak-peek and a visualization of our findings, where we show
  that after weight distribution alignment we can find a permutation map $P$ of the solution $q_{1}$ such that it can be
  linearly connected through high density regions to $q_{0}$.
\end{minipage}\hfill
\begin{minipage}{.38\textwidth}
  \centering
  \includegraphics[width=.9\textwidth]{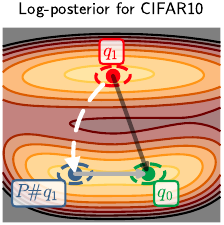}
  \\ \captionof{figure}{\textbf{Permutations in multi-modal posterior.} Log-posterior for MLP/CIFAR10, showing the two solutions ($q_{0}$ and $q_{1}$) for which we can find a permutation map such that $P_{\#}q_{1}$ can be linearly connected to $q_{0}$ with low barrier (brighter regions).}
  \label{fig:posterior_cifar}
\end{minipage}

  \section{Preliminaries on Bayesian deep learning}
In this section, we review some basic notations on \glspl{BNN} and we review \gls{SVI}, which is the main approximation
method that we analyze in this paper.
Let's consider a generic \gls{MLP} with $L$ layers, where the output of the $l$-th layer $\mbf_{l}(\mbtheta_{l}, \mbx)$ is
a vector-valued function of the previous layer output $\mbf_{l-1}$ as follows,
\begin{align}
  \mbf_{l}(\mbtheta_{l}, \mbx) = \mbW_{l}a(\mbf_{l-1}(\mbtheta_{l-1}, \mbx)) + \mbb_{l} %
\end{align}
where $a(\cdot)$ is a non-linearity, $\mbW_{l}$ is a $D_{l}\times D_{l-1}$ weight matrix and $\mbb_{l}$ the corresponding
bias vector. We shall refer to the parameters of the layer $l$ as $\mbtheta_{l}=\{\mbb_{l},\mbW_{l}\}$, and the union of
all trainable parameters as $\mbtheta = \{\mbtheta_{l}\}_{l=1}^{L}$.

The objective of using Bayesian inference on deep neural networks \citep{Mackay1994,MacKay1991} involves inferring a
posterior distribution over the parameters of the neural network given the available dataset
$\{\mbX, \mbY\} = \{(\mbx_{i}, \mby_{i})\}_{i=1}^{N}$. This requires choosing a likelihood and a prior \citep{Neal1994,Neal1994a, Tran2022}:
\begin{equation}
  p(\mbtheta\g\mbY,\mbX) = Z^{-1}p(\mbY\g \mbtheta, \mbX)p(\mbtheta)
\end{equation}
where the normalization constant $Z$ is the marginal likelihood $p(\mbY\g\mbX)$. As usually done, we assume that the likelihood
factorizes over observations, i.e. $p(\mbY\g \mbtheta, \mbX)=\prod_{i=1}^{N}p(\mby_{i}\g \mbtheta, \mbx_{i})$.

Bayesian deep learning is intractable due to the non-conjugacy likelihood-prior and thus we don't have access to closed form
solutions.
\Gls{VI} is a common technique to handle intractable Bayesian neural networks \citep{Blei2016,Hoffman2013,Graves2011,Jordan1999}.
Let $\cP(\bbR^{d})$ be the space of probability measures on $\bbR^{d}$;
\gls{VI} reframes the inference problem into an optimization one, commonly by introducing a parameterized distribution
$q(\mbtheta)\in\cP(\bbR^{d})$ which is optimized to minimize the \gls{KL} divergence with respect to the true posterior $p
(\mbtheta\g\mbY,\mbX)$. In practice, this involves the maximization of the \gls{ELBO} defined as
\begin{align}
  \label{eq:elbo}\Lelbo(q) \eqdef \int \log p(\mbY\g \mbtheta, \mbX)\dd q(\mbtheta) - \KL{q(\mbtheta)}{p(\mbtheta)}
\end{align}
whose gradients can be unbiasedly estimated with mini-batches of data \citep{Hoffman2013} and the reparameterization trick
\citep{Kingma14,Kingma2015a}.
Despite its simple formulation, the optimization of the \gls{ELBO} hides several challenges, like the initialization of
the variational parameters \citep{Rossi2018}, the effects of over-parameterization on the quality of the approximation
\citep{Rossi2020,Huix2022,Kurle2022}. Here we are interested in how different solutions to \cref{eq:elbo} relate to each
other in terms of loss barrier, which we will define formally in the following section.

\section{Loss barriers}
In the context of Bayesian inference, we are interested in the loss computed by marginalization of the model parameters
with respect to (an approximation of) the posterior. %
As such, we use the predictive likelihood, a proper scoring method for probabilistic models \citep{Rasmussen2005},
defined as
\begin{align}
  \log p(\mby^{\star}\g\mbx^{\star}) & = \log \int p(\mby^{\star}\g \mbtheta, \mbx^{\star}) p(\mbtheta\g\mbY,\mbX)\dd\mbtheta \approx \log \int p(\mby^{\star}\g \mbtheta, \mbx^{\star}) \dd q(\mbtheta) %
\end{align}
where $q(\mbtheta)$ is an approximation of the true posterior (parametric or otherwise), $\{\mbx^{\star},\mby^{\star}\}$
are respectively {the input and its corresponding label a data point under evaluation}. To keep the notation uncluttered
for the remaining of the paper, we write the predictive likelihood computed for a set of points $\{\mbx^{\star}_{i},\mby^{\star}
_{i}\}_{i=1}^{N}$ as a functional $\cL:\cP(\bbR^{d})\rightarrow\bbR$, defined as
\begin{align}
  \cL(q) \eqdef \sum_{i=1}^{N}\log \int p(\mby^{\star}_{i}\g \mbtheta, \mbx^{\star}_{i}) \dd q(\mbtheta) %
\end{align}
Let's assume two models trained with \gls{VI} with two different initializations, random seeds, and batch ordering.
Variational inference in the classic inverse sense $\KL{q}{p}$ is mode seeking, thus we expect the two runs to converge to
different solutions, say $q_{0}$ and $q_{1}$. To test the loss barrier as we interpolate between the two solutions we need
to decide on the interpolation rule. We decide to interpolate the solutions following the Wasserstein geodesics between
$q_{0}$ and $q_{1}$.
First, let's start with a few definitions. Let $q\in \cP(\bbR^{d})$ be a probability measure on $\bbR^{d}$ and $T:\bbR^{d}
\rightarrow\bbR^{d}$ a measurable map; we denote $T_{\#}q$ the \textit{push-forward measure} of $q$ through $T$. Now we
can introduce the \textit{Wasserstein geodesics}, as follows.
\begin{definition}
  The Wasserstein geodesics between $q_{0}$ and $q_{1}$ is defined as the path
  \begin{align}
    \label{eq:wasserstein_geodesics}q_{\tau}= \left((1-\tau)\mathrm{Id}+ \tau T_{q_0}^{q_1}\right)_{\#}q_{0}, \qquad \tau \in [0, 1]
  \end{align}
  where $\mathrm{Id}$ is the identity map and $T_{q_0}^{q_1}$ is the \textit{optimal transport map} between $q_{0}$ and
  $q_{1}$, which for Brenier's theorem \citep{Brenier1999}, is unique.
\end{definition}
While we could interpolate using a mixture of the two solutions, we argue that this choice is trivial and does not fully
give us a picture of the underlying loss landscape. Indeed, \cref{eq:wasserstein_geodesics} is fundamentally different
from a naive mixture path $\tilde{q}_{\tau}=(1-\tau)q_{0}+ \tau q_{1}$.
In case of Gaussian distributions, when $q_{0}=\cN(\mbm_{0}, \mbS_{0})$ and $q_{1}=\cN(\mbm_{1}, \mbS_{1})$, $q_{\tau}$ is
Gaussian as well \citep{Takatsu2011} with mean and covariance computed as follows:
\begin{align}
  \mbm_{\tau} & = (1-\tau)\mbm_{1}+ \tau\mbm_{2}\nonumber                                                                                       \\
  \mbS_{\tau} & = \mbS_{1}^{-1/2}\left( (1-\tau)\mbS_{1}+ \tau\left(\mbS_{1}^{1/2}\mbS_{2}\mbS_{1}^{1/2}\right)^{1/2}\right)^{2}\mbS_{1}^{-1/2}
\end{align}
which simplifies even further when the covariances are diagonal.

Now, we can define convexity along Wasserstein geodesics \citep{Ambrosio2008} as follows.
\begin{definition}
  Let $\cL:\cP(\bbR^{d})\rightarrow \bbR$, $\cL$ is $\lambda$ geodesics convex with $\lambda >0$ if for any
  $q_{0},q_{1}\in\cP(\bbR^{d})$ it holds that
  \begin{align}
    \cL(q_{\tau}) \leq (1-\tau)\cL(q_{0})+\tau\cL(q_{1})-\frac{\lambda \tau(1-\tau)}{2}\cW_{2}^{2}(q_{0},q_{1})
  \end{align}
  where $\cW_{2}^{2}(q_{0},q_{1})$ is the Wasserstein distance defined as \citep{villani2008optimal,Kantorovich1942,Kantorovich1948}
  \begin{align}
    \cW_{2}^{2}(q_{1}, q_{0}) = \inf_{\gamma\in\Pi(q_1, q_0)}\int \norm{\mbtheta_1-\mbtheta_0}_{2}^{2}\dd\gamma(\mbtheta_{1}, \mbtheta_{0})
  \end{align}
  with $\Pi(\cdot,\cdot)$ being the space of measure with $q_{0}$ and $q_{1}$ as marginals.
\end{definition}

While mathematically proving the geodesics convexity of the predictive likelihood for arbitrary architectures and densities
is currently beyond the scope of this work, we can empirically define a proxy using the \textit{functional loss barrier},
defined as follows.
\begin{definition}
  The functional loss barrier along the Wasserstein geodesics from $q_{0}$ and $q_{1}$ is defined as the highest difference
  between the marginal loss computed when interpolating two solutions $q_{0}$ and $q_{1}$ and the linear interpolation of
  the loss at $q_{0}$ and $q_{1}$:
  \begin{align}
    \label{eq:barrier}\cB(q_{0}, q_{1}) = \max_{\tau}\cL(q_{\tau}) - \left( (1-\tau)\cL(q_{0}) + \tau\cL(q_{1}) \right)
  \end{align}
  where $q_{\tau}$ follows the definition in \cref{eq:wasserstein_geodesics}.
\end{definition}
This definition is a more general than the ones in \citep{Ainsworth2023,Entezari2022,Frankle2020a} but we can recover
\citep{Entezari2022} by assuming delta posteriors $q_{i}=\delta(\mbtheta-\mbtheta_{i})$ and we can further recover \citep{Ainsworth2023,Frankle2020a}
by also assuming $\cL(q_{0}) = \cL(q_{1})$.

\begin{minipage}{.5\textwidth}
  \paragraph{A comment on mixtures.}
  In previous paragraphs, we argued that the mixture of distributions is not sufficient to capture the underlying complex
  geometry of the posterior. Now, we want to better illustrate this choice with a simple example. In
  \cref{fig:mixture_vs_ot} we plot the test likelihood with two interpolation strategies between two solutions ({MLP} on
  CIFAR10): the Wasserstein geodesics and the mixture. With mixtures, we see that the likelihood is pretty much constant
  during the interpolation, but this is very miss-leading: we don't see barriers not because they don't exist, but
  because the mixture simply re-weights the distributions, without continuously transporting mass in the parameter space.
\end{minipage}%
\hfill
\begin{minipage}{.45\textwidth}
  \centering
  \includegraphics[width=\textwidth]{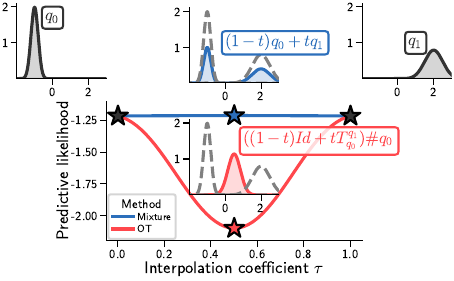}
  \captionof{figure}{\textbf{Wasserstein geodesics and mixtures.} Test likelihood for mixture and the Wasserstein geodesics interpolation. Solutions are MLPs trained on CIFAR10.}
  \label{fig:mixture_vs_ot}
\end{minipage}

  \section{Aligning distributions by looking for permutation symmetries}
\label{sec:method}

In this section, we formalize the algorithm that aligns the solutions of Bayesian inference through permutation
symmetries of weight matrices and biases.
Let $\bbS(d)$ be the set of valid $d\times d$ permutation matrices. Given a generic distribution $q(\mbtheta)$, we can apply
a permutation matrix $\mbP\in\bbS(D_{l})$ to a hidden layer output at layer $l$, and if we define $\mbtheta^{\prime}$ to
be equivalent to $\mbtheta$ with the exception of
\begin{align}
  \mbW_{l}^{\prime}= \mbP\mbW_{l}, \quad \mbb_{l}^{\prime}=\mbP\mbb_{l}, \quad \mbW_{l+1}^{\prime}=\mbW_{l+1}\mbP^{\top}\,,
\end{align}
then $P_{\#}q$ is the equivalent push-forward distribution for $\mbtheta^{\prime}$, where $P$ is the associated permutation
map.
Let us define the distribution over the functional output of the model as
\begin{align}
  q(\mbf(\mbtheta,\cdot)) = \int \delta\left(\mbf(\mbtheta,\cdot) - \mbf(\widehat\mbtheta,\cdot)\right)\dd q(\widehat\mbtheta) \,, %
\end{align}
and, equivalently, the distribution on the function using the permuted parameters as
\begin{align}
  q(\mbf(\mbtheta^{\prime},\cdot)) = \int \delta\left(\mbf(\mbtheta^{\prime},\cdot) - \mbf(\widehat\mbtheta^{\prime},\cdot)\right)\dd P_{\#}q(\widehat\mbtheta^{\prime})\,, %
\end{align}
where in both cases $\delta(\cdot)$ is the Dirac function. Then, it is simple to verify that the two models are
functionally equivalent for any inputs,
\begin{align}
  \label{eq:functional_equivalence}q(\mbf(\mbtheta, \cdot)) = q(\mbf(\mbtheta^{\prime},\cdot))\,.
\end{align}
This implies that for any weight-space distribution $q$, there exists a class of functionally equivalent solutions
$P_{\#}q$, in the sense of \cref{eq:functional_equivalence}.
These same considerations can be easily extended to other layers, by considering multiple permutation matrices $\mbP_{l}$.
For our analysis, given two solutions $q_{0}$ and $q_{1}$ we are interested in finding the permuted distribution $P_{\#}q
_{1}$, functionally equivalent to $q_{1}$, in such a way that once interpolating using \cref{eq:wasserstein_geodesics} we
observe similar performance to $q_{0}$ and $q_{1}$. Formally, we can write
\begin{align}
  \label{eq:permutation_as_optimization}\argmin_{P}\cD(q_{1}(\mbf(\mbtheta^{\prime}, \cdot)), q_{0}(\mbf(\mbtheta, \cdot))) = \argmin_{P}\cD(P_{\#}q_{1}(\mbtheta), q_{0}(\mbtheta))\,,
\end{align}
where $\cD$ is a generic measure of discrepancy.\footnotemark \footnotetext{To be formally correct, the l.h.s. is a
discrepancy defined on stochastic processes while the r.h.s. is defined on random vectors. Under mild assumptions on the
distribution on the parameters and the architectures, $\cD$ is well defined in both cases \citep[see e.g.,][]{Polyanskiy2017}.}

\subsection{Problem setup for permutation of vectors}
We start from a single vector of parameters, disregarding for the moment the functional equivalence constrain. We will
extend these results to matrices and multiple layers later. In practice considering Gaussian distributions, we know that
if $q = \cN(\mbm, \mbS)$, then $P_{\#}q = \cN(\mbP\mbm, \mbP\mbS\mbP^{\top})$ and if $q = \cN(\mbm, \diag(\mbs^{2}))$ then
$P_{\#}q = \cN(\mbP\mbm, \diag(\mbP\mbs^{2}))$ \citep{Bishop2006}. With the \gls{KL} divergence $\KL{P_\#q_1}{q_0}$ %
it's easy to verify that it leads to just a distance between means, disregarding any covariance information.
While certainly this represents a valid choice, we argue that we can find a better solution by using the Wasserstein
distance.
For Gaussian measures, the Wasserstein distance has analytic solution:
\begin{align}
  \cW_{2}^{2}({q_1},{q_0}) & = \norm{\mbm_0 - \mbm_1}^{2}_{2}+ \Tr(\mbS_{1}+ \mbS_{0}- 2\left(\mbS_{1}^{1/2}\mbS_{0}\mbS_{1}^{1/2}\right)^{1/2}) =\nonumber \\
                           & = \norm{\mbm_0 - \mbm_1}^{2}_{2}+ \norm{\mbS_0 - \mbS_1}^{2}_{F}\,,
\end{align}
where $\norm{\cdot}_{F}$ denotes the Frobenius norm, $\norm{\mbA}=\sum_{ij}{a_{ij}}^{2}$, and where the second line is valid
only if the covariances commute ($\mbS_{1}\mbS_{0}=\mbS_{0}\mbS_{1}$). In our case, then, we can simplify as follows:
\begin{align}
  \cW_{2}^{2}({P_\#q_1},{q_0}) & = \norm{\mbm_0 - \mbP\mbm_1}^{2}_{2}+ \norm{\mbs_0 - \mbP\mbs_1}_{2}^{2}\,.
\end{align}
To summarize, the problem now can be written as:
\begin{align}
  \label{eq:lap}\argmin_{\mbP\in\bbS(d)}\norm{\mbm_0 - \mbP\mbm_1}^{2}_{2}+ \norm{\mbs_0 - \mbP\mbs_1}_{2}^{2}= \argmax_{\mbP\in\bbS(d)}\innerproduct{\mbP}{\mbm_0\mbm_1^\top + \mbs_0\mbs_1^\top}_{F}\,,
\end{align}
where the expression $\innerproduct{\mbA}{\mbB}_{F}$ is the Frobenius inner product,
$\innerproduct{\mbA}{\mbB}_{F}=\sum_{ij}A_{ij}B_{ij}$. Note that the r.h.s. of \cref{eq:lap} is a valid instantiation of
the \gls{LAP} \citep{Bertsekas1998}, which can be solved in polynomial time.

\subsection{From vectors to neural network parameters}
Finally, we need to take into account that we have multiple layers and weight matrices, and that we are trying to find functionally
equivalent solutions. For this, we decide to explicitly change our main objective by enforcing the functional equivalence
constraint as follows:
\begin{align*}
  \argmin_{\{P_i\}}\cW_{2}^{2}\left({P_{1\#}q_1^{(1)}},{q_0^{(1)}}\right) + \cW_{2}^{2}\left({\left(P_2\circ P^\top_{1}\right)_\#q_1^{(2)}},{q_0^{(2)}}\right) + \dots + \cW_{2}^{2}\left({\left(P^\top_{L-1}\right)_\#q_1^{(L)}},{q_0^{(L)}}\right) \,,
\end{align*}
where the notation $\left(P_{l}\circ P^{\top}_{l-1}\right)$ represents the composition of the two permutation maps
applied to rows and columns of the random weight matrices. More conveniently, this can be rewritten in terms of means
and standard deviations. To leave the notation uncluttered, let's collect the means and the standard deviations for the
layer $l$ in $\mbM^{(l)}$ and $\mbS^{(l)}$, which are now both $D_{l}\times D_{l-1}$ matrices, so that
$q^{(l)}=\prod_{ij}\cN(M^{(l)}_{ij}, S^{(l)}_{ij})$. Now we can write,
\begin{align}
  \argmax_{\left\{ \mbP_i \right\}_{i=1}^L}\; & \innerproduct{\mbM_0^{(1)}}{\mbP_1 \mbM_1^{(1)}}_{F}+ \innerproduct{\mbS_0^{(1)}}{\mbP_1 \mbS_1^{(1)}}_{F}+ \nonumber \innerproduct{\mbM_0^{(2)}}{\mbP_2 \mbM_1^{(2)}\mbP_1^\top}_{F}+ \innerproduct{\mbS_0^{(2)}}{\mbP_2 \mbS_1^{(2)}\mbP_1^\top}_{F}+ \nonumber \\
                                              & + \dots + \nonumber \innerproduct{\mbM_0^{(L)}}{\mbM_1^{(L)}\mbP_{L-1}^\top}_{F}+ \innerproduct{\mbS_0^{(L)}}{\mbS_{1}^{(L)}\mbP_{L-1}^\top}_{F}\,.                                                                                                               %
\end{align}
This optimization problem is more challenging than the one presented in \cref{eq:lap}: we are interested in finding permutation
matrices to be applied concurrently to rows and columns of both means and standard deviations. This class of problems, also
known as \gls{SOLAP}, is NP-hard and no polynomial-time solutions exist. For this reason, we propose to use the setup in
\citet{Ainsworth2023} by extending it to our problem.
In particular, by fixing all matrices with the exception of ${\mbP_{l}}$, we observe that also in our case the problem can
be reduced to a classic \gls{LAP}.
\begin{align}
  \label{eq:align_vi}\argmax_{\mbP_l} & \innerproduct{\mbM_0^{(l)}}{\mbP_l \mbM_1^{(l)} \mbP_{l-1}^\top }_{F}+ \innerproduct{\mbM_0^{(l+1)}}{\mbP_{(l+1)} \mbM_1^{(l+1)} \mbP_l^\top}_{F}+ \nonumber        \\
                                      & \innerproduct{\mbS_0^{(l)}}{\mbP_l \mbS_1^{(l)} \mbP_{l-1}^\top }_{F}+ \innerproduct{\mbS_0^{(l+1)}}{\mbP_{(l+1)} \mbS_1^{(l+1)} \mbP_l^\top}_{F}= \nonumber        \\
  =\argmax_{\mbP_l}                   & \Big\langle \mbP_{l}\Big\vert \mbM_{0}^{(l)}\mbP_{l-1}\left(\mbM_{1}^{(l)}\right)^{\top}+ \left(\mbM_{0}^{(l+1)}\right)^{\top}\mbP_{l+1}\mbM_{1}^{(l+1)}+ \nonumber \\
                                      & \qquad\mbS_{0}^{(l)}\mbP_{l-1}\left(\mbS_{1}^{(l)}\right)^{\top}+ \left(\mbS_{0}^{(l+1)}\right)^{\top}\mbP_{l+1}\mbS_{1}^{(l+1)}\Big\rangle_{F}\,.                  %
\end{align}
As discussed in \citep{Ainsworth2023}, going through each layer, and greedily selecting its best $\mbP_{l}$, leads to a coordinate
descent algorithm which guarantees to end in finite time. We present a pseudo-code in \cref{alg:align_vi}.

\begin{algorithm}
  \small
  \caption{Algorithm to align variational inference solutions}
  \label{alg:align_vi}%
  \KwData{Variational inference solutions $q_{0}$ and $q_{1}$}%
  \KwResult{Permutation matrices $\mbP_{i}$}%
  $\mbP_{i}\gets \mbI$, $\forall i\in\{1,\dots,L\}$\;%
  \While{not converged}{%
  \For{$i \in \text{RandomPerm}(1, \dots, L-1)$}{

  $\mbP_{i}= \argmax \innerproduct{\mbP_i}{\star}$, where $\star$ is the r.h.s. of \cref{eq:align_vi} %
  } }
\end{algorithm}

  \begin{figure}[t]
  \centering
  \begin{subfigure}
    [t]{0.33\textwidth}
    \centering
    \includegraphics[width=.99\textwidth]{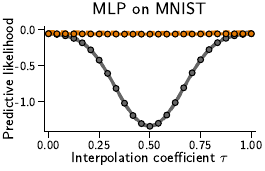}
  \end{subfigure}%
  \begin{subfigure}
    [t]{0.33\textwidth}
    \centering
    \includegraphics[width=.99\textwidth]{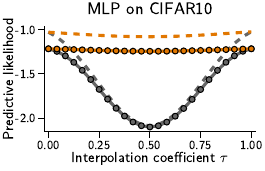}
  \end{subfigure}%
  \begin{subfigure}
    [t]{0.33\textwidth}
    \centering
    \includegraphics[width=.99\textwidth]{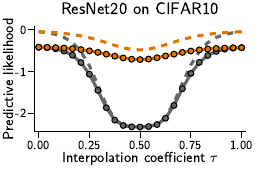}
  \end{subfigure}\\
  \setlength{\tabcolsep}{2pt}
\scriptsize

\begin{tabular}{clclclcl}
  \toprule                                                                                                                                    %
  \tikz[baseline=-1ex]{\draw[color=black!60, fill=black!60, line width=1.4pt, dashed, dash pattern=on 5pt off 2pt] plot[] (0.0,0)--+(.7,0);} & \textsf{VI (Train)} & \tikz[baseline=-1ex]{\draw[color=xkcdPumpkin, fill=xkcdPumpkin, line width=1.4pt, dashed, dash pattern=on 5pt off 2pt] plot[] (0.0,0)--+(.7,0);} & \textsf{VI with distr. alignment (Train)} & \tikz[baseline=-1ex]{\draw[color=black!60, fill=black!60, line width=1.4pt] plot[] (0.0,0)--+(-.3,0)--+(.3,0); \draw[color=black, fill=black!60, mark size=3pt, mark=*, line width=.5pt] plot[] (0.0,0)--+(-.01,0)--+(.01,0);} & \textsf{VI (Test)} & \tikz[baseline=-1ex]{\draw[color=xkcdPumpkin, fill=xkcdPumpkin, line width=1.4pt] plot[] (0.0,0)--+(-.3,0)--+(.3,0); \draw[color=black, fill=xkcdPumpkin, mark size=3pt, mark=*, line width=.5pt] plot[] (0.0,0)--+(-.01,0)--+(.01,0);} & \textsf{VI with distr. alignment (Test)} \\
  \bottomrule
\end{tabular}
  \caption{\textbf{{Zero barrier solutions.}} Comparison of loss barriers for standard \gls{VI} (gray) and \gls{VI} with
  alignment (orange).
  While loss barriers always appear between two solutions in the standard \gls{VI} approach, in the case of \gls{VI} with
  alignment there is no noticeable loss barrier for \glspl{MLP}
  and a nearly-zero loss barrier for ResNet20.
  }
  \label{fig:zero-barrier_interpolation}
\end{figure}
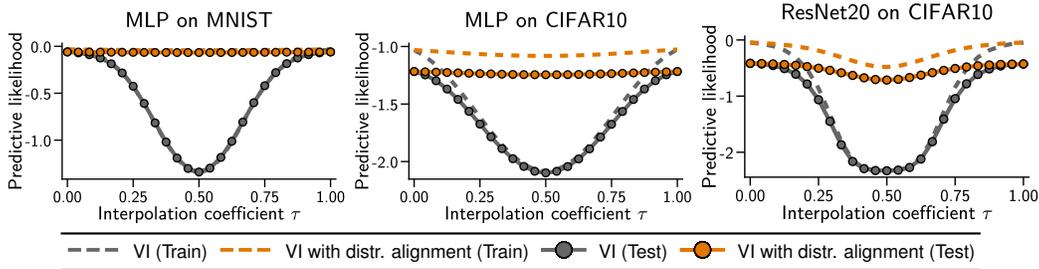

\section{Experiments}
\label{sec:experiments}

Now, we present some supporting evidence to \cref{claim:lmc}. We start by training two replicas of \glspl{BNN} with variational
inference (we refer to the Appendix for additional details on the experimental setup). We then compute the marginalized
barrier as $\cB(q_{0}, q_{1}) = \max_{\tau}\cL(q_{\tau}) - \left( (1-\tau)\cL(q_{0}) + \tau \cL(q_{1}) \right)$ where
$\cL(\cdot)$ is the predictive likelihood and $\tau\in[0, 1]$, from which we take 25 evenly distributed points. In
particular, we seek to understand what happens to the \gls{VI} solutions first for the naive interpolation from $q_{0}$
and $q_{1}$, and then for the interpolation after aligning $q_{0}$ and $P_{\#}q_{1}$.
We experiment with \glspl{MLP} with three layers and ResNet20 \citep{He2016} with various widths on MNIST \citep{mnist},
Fashion-MNIST \citep{fmnist} and CIFAR10 \citep{cifar}. All models are trained without data augmentation \citep{Wenzel2020}
and with \gls{FRN} layers instead of BatchNorm. Finally, we set the prior to be Gaussian $\cN(\mbzero, \alpha^{2}\mbI)$,
with the flexibility of choosing the variance.

\subsection{Low-barrier interpolations}

\cref{fig:zero-barrier_interpolation} shows the results with and without alignment. We see that regardless of the
dataset and the model used, the performance degrades significantly when we move between the two solutions with the naive
interpolation, showing the existence of barriers in the predictive likelihood for Gaussian \gls{VI} solutions. However, with
the alignment proposed in \cref{sec:method} and \cref{alg:align_vi}, we recover zero barrier solutions for \glspl{MLP} on
both MNIST and CIFAR10, and nearly-zero barrier for ResNet20 on CIFAR10. This holds both for the train and test splits,
with quantifiably smaller barriers in the test set.

\begin{minipage}{.46\textwidth}
  In \cref{fig:barrier_vs_width} we study the effect of the width of a neural network in relation to the loss barrier by
  taking an \gls{MLP} and a ResNet20 with an increasing number of hidden features. We see that wider models generally provide
  lower barriers: for \glspl{MLP} this holds true with and without alignment, while for the ResNet20 this is happening
  only after alignment. This extends some previous analysis done on loss-optimized networks. Specifically, \citet{Entezari2022}
  show that barriers seem to have a double descent trend, while \citet{Ainsworth2023} discuss that low barrier solutions
  after accounting for symmetries are easier to find in wider networks.
\end{minipage}\hfill
\begin{minipage}{.5\textwidth}
  \centering
  \includegraphics[width=.5\textwidth]{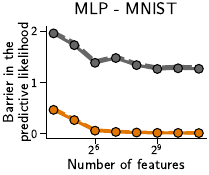}%
  \includegraphics[width=.5\textwidth]{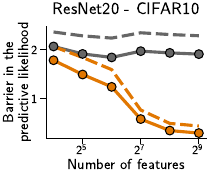}
  \\
  \setlength{\tabcolsep}{2pt}
\scriptsize

\begin{tabular}{clclclcl}
  \toprule                                                                                                                                          %
  \tikz[baseline=-1ex]{\draw[color=black!60, fill=black!60, line width=1.4pt, dashed, dash pattern=on 5pt off 2pt] plot[] (0.0,0)--+(.7,0);}       & \textsf{VI (Train)}             & \tikz[baseline=-1ex]{\draw[color=black!60, fill=black!60, line width=1.4pt] plot[] (0.0,0)--+(-.3,0)--+(.3,0); \draw[color=black, fill=black!60, mark size=3pt, mark=*, line width=.5pt] plot[] (0.0,0)--+(-.01,0)--+(.01,0);}          & \textsf{VI (Test)}             \\
  \tikz[baseline=-1ex]{\draw[color=xkcdPumpkin, fill=xkcdPumpkin, line width=1.4pt, dashed, dash pattern=on 5pt off 2pt] plot[] (0.0,0)--+(.7,0);} & \textsf{VI with align. (Train)} & \tikz[baseline=-1ex]{\draw[color=xkcdPumpkin, fill=xkcdPumpkin, line width=1.4pt] plot[] (0.0,0)--+(-.3,0)--+(.3,0); \draw[color=black, fill=xkcdPumpkin, mark size=3pt, mark=*, line width=.5pt] plot[] (0.0,0)--+(-.01,0)--+(.01,0);} & \textsf{VI with align. (Test)} \\
  \bottomrule
\end{tabular}
  \captionof{figure}{\textbf{Effect of width.} After distribution alignment, wider models exhibit lower likelihood barrier.}%
  \label{fig:barrier_vs_width}
\end{minipage}

We speculate that this might be due to the limiting behavior of Bayesian neural networks, which makes the posterior landscape
Gaussian-like \citep{Jacot2018,He2020}. %
While this does not fully explain the phenomenon observed, the existing connections between \glspl{BNN} and non-parametric
models, like \glspl{GP} \citep{Rasmussen2005} and \glspl{DGP} \citep{Damianou13,Cutajar17,Matthew2018,Salimbeni17}, can
provide additional insights on the role of symmetries in weight space \citep{Pleiss2021}.

\begin{figure}[t]
  \begin{subfigure}
    []{.4\textwidth}
    \centering
    \includegraphics[width=\textwidth]{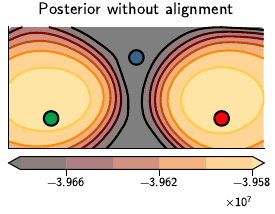}
    \\[-3ex] $(a)$
  \end{subfigure}%
  \begin{subfigure}
    []{.4\textwidth}
    \centering
    \includegraphics[width=\textwidth]{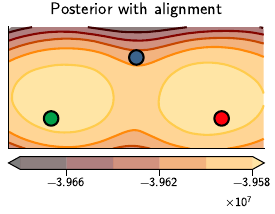}
    \\[-3ex] $(b)$
  \end{subfigure}%
  \begin{subfigure}
    []{.2\textwidth}
    \setlength{\tabcolsep}{2.5pt}
\scriptsize \sffamily
\begin{tabular}{cl|cl}
  \toprule                                                                                                                                     %
  $(a)$                                                                                                                                       &            & $(b)$                                                                                                                                        \\ %
  \midrule                                                                                                                                     %
  \tikz[baseline=-0.5ex]{\draw[color=black, fill=xkcdEmerald, mark size=3pt, mark=*, line width=.5pt] plot[] (0.0,0)--+(-.01,0)--+(.01,0);}   & $q_{0}$    & \tikz[baseline=-0.5ex]{\draw[color=black, fill=xkcdEmerald, mark size=3pt, mark=*, line width=.5pt] plot[] (0.0,0)--+(-.01,0)--+(.01,0);}   & $q_{0}$          \\
  \tikz[baseline=-0.5ex]{\draw[color=black, fill=xkcdDenim, mark size=3pt, mark=*, line width=.5pt] plot[] (0.0,0)--+(-.01,0)--+(.01,0);}     & $q_{\tau}$ & \tikz[baseline=-0.5ex]{\draw[color=black, fill=xkcdDenim, mark size=3pt, mark=*, line width=.5pt] plot[] (0.0,0)--+(-.01,0)--+(.01,0);}     & $P_{\#}q_{\tau}$ \\
  \tikz[baseline=-0.5ex]{\draw[color=black, fill=xkcdBrightRed, mark size=3pt, mark=*, line width=.5pt] plot[] (0.0,0)--+(-.01,0)--+(.01,0);} & $q_{1}$    & \tikz[baseline=-0.5ex]{\draw[color=black, fill=xkcdBrightRed, mark size=3pt, mark=*, line width=.5pt] plot[] (0.0,0)--+(-.01,0)--+(.01,0);} & $P_{\#}q_{1}$    \\
  \bottomrule
\end{tabular}

Note: $\tau = 0.5$
  \end{subfigure}
  \caption{ \textbf{Posterior density visualization.} Analysis of the log-posterior computed for ResNet20. Samples from $q
  _{0}$ and $q_{1}$ are connected by lower density regions, while $q_{0}$ and $P_{\#}q_{1}$ are not.}
  \label{fig:log_posterior_2d}
\end{figure}

Finally, as an additional check, we analyze the log-posterior with and without alignment by projecting the density into two
dimensional slices, following the setup in \citep{Izmailov21,Garipov2018}. We study the two dimensional subspace of the
parameter space supported by the hyperplane %
$H = \{\mbtheta \in\bbR^{d}\g \mbtheta = a\mbtheta_{a}+ b\mbtheta_{b}+ (1 - a - b)\mbtheta_{c}\}\,,$,
where $a,b\in\bbR$ and $\mbtheta_{a}$, $\mbtheta_{b}$ and $\mbtheta_{c}$ are the samples either from $q_{0}$, $q_{1}$
and $q_{\tau}$ without alignment or from $q_{0}$, $P_{\#}q_{1}$ and $P_{\#}q_{\tau}$ with alignment. With this
configuration, all three samples always lie on this hyper-plane. In \cref{fig:log_posterior_2d}, we present the visualization
of ResNet20 trained on CIFAR10. We see that the samples from $q_{0}$ and $P_{\#}q_{1}$ are connected by higher density
regions than the ones between $q_{0}$ and $q_{1}$. This is in line with the results in \cref{fig:zero-barrier_interpolation},
where we see that the loss barrier is lower after alignment.

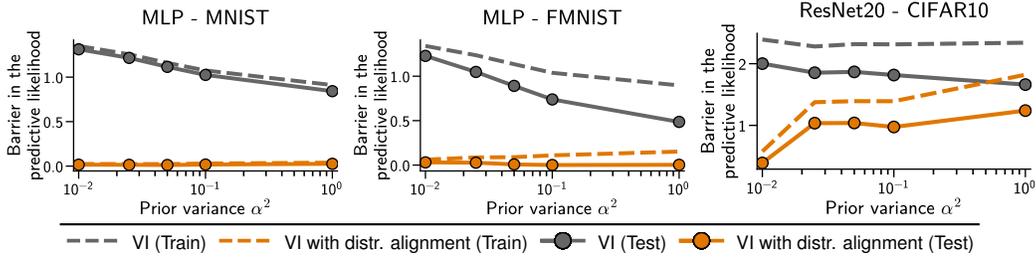
\begin{figure}[t]
  \centering
  \begin{subfigure}
    {.33\textwidth}
    \centering
    \includegraphics[width=.99\textwidth]{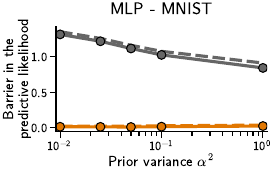}%
  \end{subfigure}%
  \begin{subfigure}
    {.33\textwidth}
    \centering
    \includegraphics[width=.99\textwidth]{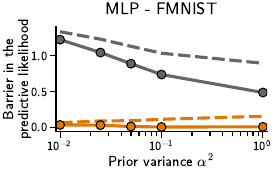}%
  \end{subfigure}%
  \begin{subfigure}
    {.33\textwidth}
    \centering
    \includegraphics[width=.99\textwidth]{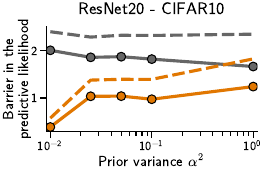}%
  \end{subfigure}\\
  \setlength{\tabcolsep}{2pt}
\scriptsize

\begin{tabular}{clclclcl}
  \toprule                                                                                                                                    %
  \tikz[baseline=-1ex]{\draw[color=black!60, fill=black!60, line width=1.4pt, dashed, dash pattern=on 5pt off 2pt] plot[] (0.0,0)--+(.7,0);} & \textsf{VI (Train)} & \tikz[baseline=-1ex]{\draw[color=xkcdPumpkin, fill=xkcdPumpkin, line width=1.4pt, dashed, dash pattern=on 5pt off 2pt] plot[] (0.0,0)--+(.7,0);} & \textsf{VI with distr. alignment (Train)} & \tikz[baseline=-1ex]{\draw[color=black!60, fill=black!60, line width=1.4pt] plot[] (0.0,0)--+(-.3,0)--+(.3,0); \draw[color=black, fill=black!60, mark size=3pt, mark=*, line width=.5pt] plot[] (0.0,0)--+(-.01,0)--+(.01,0);} & \textsf{VI (Test)} & \tikz[baseline=-1ex]{\draw[color=xkcdPumpkin, fill=xkcdPumpkin, line width=1.4pt] plot[] (0.0,0)--+(-.3,0)--+(.3,0); \draw[color=black, fill=xkcdPumpkin, mark size=3pt, mark=*, line width=.5pt] plot[] (0.0,0)--+(-.01,0)--+(.01,0);} & \textsf{VI with distr. alignment (Test)} \\
  \bottomrule
\end{tabular}
  \caption{\textbf{Effect of prior variance.} After distribution alignment, prior variance has low effect in finding
  zero-loss barriers, while with naive interpolation we see a decreasing trend the higher the variance is.}
  \label{fig:barrier_vs_prior}
\end{figure}
\subsection{Analyzing the effect of the prior and testing the cold posterior effect}

In all previous experiments we used a Gaussian prior $\cN(\mbzero, \alpha^{2}\mbI)$ with fixed $\alpha^{2}$; now we
study the effect of a varying prior variance $\alpha^{2}$ on the behavior of the loss barriers. We experiment this on a \gls{MLP}
trained on MNIST and Fashion-MNIST and on a ResNet20 (width x8) on CIFAR10. We report the results in \cref{fig:barrier_vs_prior}.
We can appreciate two behaviors: with alignment, there is no measurable effect of using different variances in finding
zero-barrier solutions; on the contrary, without alignment we see that naive \gls{VI} solutions are easier to interpolate
with lower barrier when the prior is more diffused. At the same time, we see that higher variances produce bigger gaps between
train barriers and test barriers. We speculate that this is due to overfitting happening with more relaxed priors, which
makes low-barrier (but low-likelihood) solutions easier to find.

\begin{figure}[t]
  \centering
  \begin{subfigure}
    {.33\textwidth}
    \centering
    \includegraphics[width=.99\textwidth]{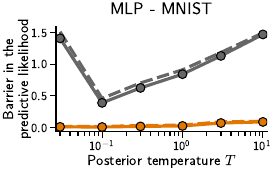}%
  \end{subfigure}%
  \begin{subfigure}
    {.33\textwidth}
    \centering
    \includegraphics[width=.99\textwidth]{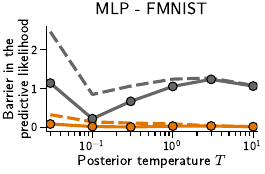}%
  \end{subfigure}%
  \begin{subfigure}
    {.33\textwidth}
    \centering
    \includegraphics[width=.99\textwidth]{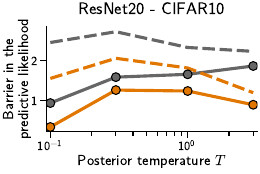}%
  \end{subfigure}\\
  \setlength{\tabcolsep}{2pt}
\scriptsize

\begin{tabular}{clclclcl}
  \toprule                                                                                                                                    %
  \tikz[baseline=-1ex]{\draw[color=black!60, fill=black!60, line width=1.4pt, dashed, dash pattern=on 5pt off 2pt] plot[] (0.0,0)--+(.7,0);} & \textsf{VI (Train)} & \tikz[baseline=-1ex]{\draw[color=xkcdPumpkin, fill=xkcdPumpkin, line width=1.4pt, dashed, dash pattern=on 5pt off 2pt] plot[] (0.0,0)--+(.7,0);} & \textsf{VI with distr. alignment (Train)} & \tikz[baseline=-1ex]{\draw[color=black!60, fill=black!60, line width=1.4pt] plot[] (0.0,0)--+(-.3,0)--+(.3,0); \draw[color=black, fill=black!60, mark size=3pt, mark=*, line width=.5pt] plot[] (0.0,0)--+(-.01,0)--+(.01,0);} & \textsf{VI (Test)} & \tikz[baseline=-1ex]{\draw[color=xkcdPumpkin, fill=xkcdPumpkin, line width=1.4pt] plot[] (0.0,0)--+(-.3,0)--+(.3,0); \draw[color=black, fill=xkcdPumpkin, mark size=3pt, mark=*, line width=.5pt] plot[] (0.0,0)--+(-.01,0)--+(.01,0);} & \textsf{VI with distr. alignment (Test)} \\
  \bottomrule
\end{tabular}
  \caption{\textbf{Effect of temperature.} After alignment, cold posteriors makes barriers marginally closer to zero}
  \label{fig:barrier_vs_temperature}
\end{figure}
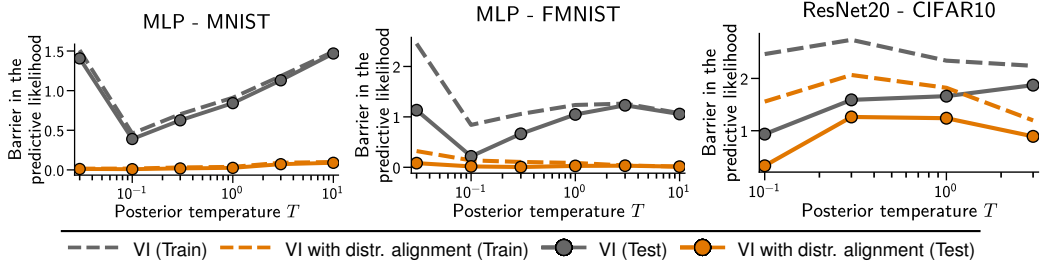

Additionally, several previous works have analyzed the effect of tempering the posterior in \glspl{BNN} \citep{Wenzel2020,ZhangLZCW20,zeno2021why,Izmailov21}.
Specifically, we are interested in the distribution
$p_{T}(\mbtheta\g\mbY)\propto \left(p(\mbY\g \mbtheta,\mbX)p(\mbtheta )\right)^{1/T}$, where $T$ is known as the temperature.
Note that starting from the above definition, we can write an equivalent \gls{ELBO} for \gls{VI} which takes into account
$T$. For $T<1$, we have cold posteriors, which are sharper than the true Bayesian posteriors, while for $T>1$ we have warm
posterior, which are more diffused. In \cref{fig:barrier_vs_temperature} we see that barriers for cold posteriors with alignment
are marginally closer to zero than for warm posteriors. Note that cold temperatures concentrate the distribution around
the \gls{MAP}, which motivates a further comparison with a non-Bayesian approach.

\subsection{Comparison with SGD solutions}

\begin{minipage}{.5\textwidth}
  Motivated by the results with cold posteriors, we also compare the behavior of barriers for \gls{VI} versus \gls{MAP}
  solutions obtained via \gls{SGD}. Note that using the \gls{MAP} solution we end up with the same setup of weight matching
  than in \citep{Ainsworth2023}. In \cref{fig:vi_vs_sgd} we report this comparison, which is carried out on ResNet20 for
  various width multipliers. For narrow models, we see that \gls{VI} with alignment and \gls{MAP} with weight matching
  both behave in a similar manner. Interestingly, as the model becomes wider, \gls{MAP} solutions achieve marginally
  lower barriers than \gls{VI}. We speculate that this might be due to the simple Gaussian parameterization for the
  approximate posterior, which doesn't completely capture the local geometry of the true posterior.
\end{minipage}\hfill
\begin{minipage}{.46\textwidth}
  \centering
  \includegraphics[width=.93\textwidth]{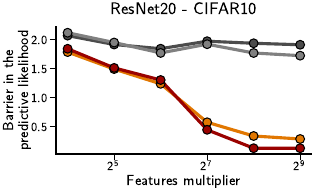}
  \\
  \definecolor{color_1}{HTML}{df2b4f}
\definecolor{color_2}{HTML}{19198c}
\definecolor{color_3}{HTML}{aa6143}
\definecolor{color_4}{HTML}{900289}
\definecolor{color_5}{HTML}{ff9a36}
\definecolor{color_6}{HTML}{009695}

\setlength{\tabcolsep}{1.8pt}
\scriptsize

\begin{tabular}{clclcl}
  \toprule                                                                                                                                                                                                                                               %
  \tikz[baseline=-1ex]{\draw[color=black!60, fill=black!60, line width=1.4pt] plot[] (0.0,0)--+(-.3,0)--+(.3,0); \draw[color=black, fill=black!60, mark size=3pt, mark=*, line width=.5pt] plot[] (0.0,0)--+(-.01,0)--+(.01,0);}                        & \textsf{VI}                & \tikz[baseline=-1ex]{\draw[color=black!40, fill=black!40, line width=1.4pt] plot[] (0.0,0)--+(-.3,0)--+(.3,0); \draw[color=black, fill=black!40, mark size=3pt, mark=*, line width=.5pt] plot[] (0.0,0)--+(-.01,0)--+(.01,0);}                        & \textsf{SGD}                      \\
  \tikz[baseline=-1ex]{\draw[color=xkcdPumpkin, fill=xkcdPumpkin, opacity=0.99, line width=1.4pt] plot[] (0.0,0)--+(-.3,0)--+(.3,0); \draw[color=black, fill=xkcdPumpkin, mark size=3pt, mark=*, line width=.5pt] plot[] (0.0,0)--+(-.01,0)--+(.01,0);} & \textsf{VI with alignment} & \tikz[baseline=-1ex]{\draw[color=xkcdDeepRed, fill=xkcdDeepRed, opacity=0.99, line width=1.4pt] plot[] (0.0,0)--+(-.3,0)--+(.3,0); \draw[color=black, fill=xkcdDeepRed, mark size=3pt, mark=*, line width=.5pt] plot[] (0.0,0)--+(-.01,0)--+(.01,0);} & \textsf{SGD with weight matching} \\
  \bottomrule
\end{tabular}
  
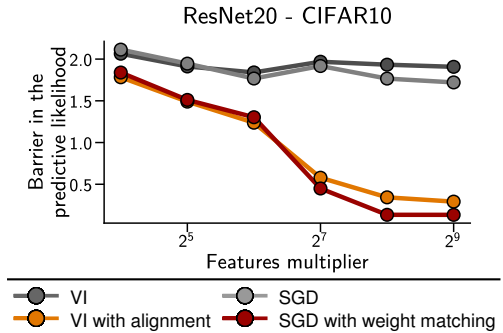
\captionof{figure}{\textbf{\gls{VI} versus \gls{SGD}.} \gls{VI} and \gls{SGD} behave equivalently after permutation alignment in narrow models. For wider models, \gls{SGD} solutions reach lower barriers than \gls{VI}.}
  \label{fig:vi_vs_sgd}
\end{minipage}

\subsection{Effect of normalization layers and data augmentation}

We conclude this section with a discussion on the effects of normalization layers and data augmentation on the loss barriers
for Bayesian neural networks.

\begin{figure}
  \centering
  \begin{minipage}[t]{.33\textwidth}
    \centering
    \includegraphics[width=.98\textwidth]{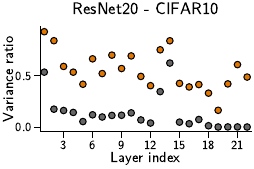}
    \setlength{\tabcolsep}{2.5pt}
\scriptsize \sffamily
\begin{tabular}{cl}
  \toprule                                                                                                                                   %
  \tikz[baseline=-0.5ex]{\draw[color=black, fill=xkcdPumpkin, mark size=3pt, mark=*, line width=.5pt] plot[] (0.0,0)--+(-.01,0)--+(.01,0);} & Interpolation after alignment \\
  \tikz[baseline=-0.5ex]{\draw[color=black, fill=black!60, mark size=3pt, mark=*, line width=.5pt] plot[] (0.0,0)--+(-.01,0)--+(.01,0);}    & Naive interpolation           \\
  \bottomrule
\end{tabular}
    \\[1ex]
    \begin{minipage}{.9\textwidth}
      \centering
      \captionof{figure}{\textbf{Variance collapse.} Variance collapse is present for the naive interpolation, but it is reduced after alignment.}
      \label{fig:variance_collapse}
    \end{minipage}
  \end{minipage}%
  \begin{minipage}[t]{.33\textwidth}
    \centering
    \includegraphics[width=.98\textwidth]{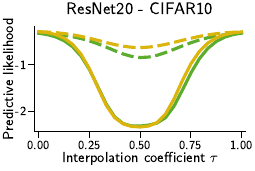}
    \setlength{\tabcolsep}{1.8pt}
\scriptsize \sffamily

\begin{tabular}{clclcl}
  \toprule                                                                                                 %
                                                                                                          & \textbf{Norm. layer} &                                                                                                                                            & \textbf{Alignment} \\
  \tikz[baseline=-1ex]{\draw[color=xkcdGrass, fill=xkcdGrass, line width=1.4pt] plot[] (0.0,0)--+(.5,0);} & FRN                  & \tikz[baseline=-1ex]{\draw[color=black!60, fill=black!40, line width=1.4pt, dashed, dash pattern=on 5pt off 2pt] plot[] (0.0,0)--+(.5,0);} & True               \\
  \tikz[baseline=-1ex]{\draw[color=xkcdGold, fill=xkcdGold, line width=1.4pt] plot[] (0.0,0)--+(.5,0);}   & LN                   & \tikz[baseline=-1ex]{\draw[color=black!60, fill=black!40, line width=1.4pt] plot[] (0.0,0)--(.5,0);}                                       & False              \\
  \bottomrule
\end{tabular}
    \\[1ex]
    \begin{minipage}{.9\textwidth}
      \centering
      \captionof{figure}{\textbf{Effect of normalization layers.} Both FRN and LN can be aligned, and LN exhibits lower barriers.}
      \label{fig:barrier_vs_norm}
    \end{minipage}
  \end{minipage}
  \begin{minipage}[t]{.33\textwidth}
    \centering
    \includegraphics[width=.98\textwidth]{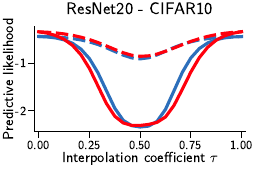}
    \setlength{\tabcolsep}{1.8pt}
\scriptsize \sffamily

\begin{tabular}{clclcl}
  \toprule                                                                                                           %
                                                                                                                    & \textbf{DA} &                                                                                                                                            & \textbf{Alignment} \\
  \tikz[baseline=-1ex]{\draw[color=xkcdBrightRed, fill=xkcdBrightRed, line width=1.4pt] plot[] (0.0,0)--+(.5,0);}   & True        & \tikz[baseline=-1ex]{\draw[color=black!60, fill=black!40, line width=1.4pt, dashed, dash pattern=on 5pt off 2pt] plot[] (0.0,0)--+(.5,0);} & True               \\
  \tikz[baseline=-1ex]{\draw[color=xkcdMediumBlue, fill=xkcdMediumBlue, line width=1.4pt] plot[] (0.0,0)--+(.5,0);} & False       & \tikz[baseline=-1ex]{\draw[color=black!60, fill=black!40, line width=1.4pt] plot[] (0.0,0)--(.5,0);}                                       & False              \\
  \bottomrule
\end{tabular}
    \\[1ex]
    \begin{minipage}{.9\textwidth}
      \centering
      \captionof{figure}{\textbf{Effect of data augmentation.} With and without DA, we are able to recover similar low barrier solutions.}
      \label{fig:barrier_vs_da}
    \end{minipage}
  \end{minipage}%
\end{figure}

Different normalization strategies can affect the overall geometry of the problem \citep{Entezari2022,Ainsworth2023}. As
discussed in \citep{Jordan2023}, interpolating with BatchNorm layers \citep{Ioffe2015} is pathological due to \textit{variance
collapse} of the feature representation in hidden layers. Additionally, batch-dependent normalization layers don't have a
clear Bayesian interpretation, since the likelihood cannot factorize. For our experiments we choose to use the \gls{FRN}
layer, as done in previous works \citep[e.g.,][]{Izmailov21}. Note that \gls{FRN} layers are invariant to permutation units
and therefore can be aligned without problems. To test the \emph{variance collapse} behavior, we analyze the variance of
activations following the instructions in \citep[\S3.1]{Jordan2023}, with the sole difference that the activations are
marginalized w.r.t. samples from the posterior. In \cref{fig:variance_collapse} we can see that there isn't a
pathological variance collapse after alignment. Finally, in \cref{fig:barrier_vs_norm} we compare another normalization
layer, the \gls{LN} \citep{Ba2016}. Note that \gls{LN} is also batch-independent, it has a clear Bayesian interpretation
and it is invariant to permutation units. Indeed, we see that both normalization layers can be aligned, and \gls{LN}
exhibits lower barriers than \gls{FRN}.

Finally, in all previous experiments we skipped data augmentation, because the random augmentations introduce stochasticity
which lacks a proper Bayesian interpretation in the formulation of the likelihood function (e.g. re-weighting of the likelihood
due to the increase of the effective sample size \citep{Osawa2019}). Additionally, data augmentation can contribute to spurious
effects difficult to disentangle (e.g., cold posterior effect \citep{Wenzel2020}). Nonetheless, during the development
of the method we didn't make an assumption on data augmentation and in \cref{fig:barrier_vs_da} we experiment both with
and without augmentation, showing that we are still able to recover similar low barrier solutions in both cases. Having
said that, we advocate caution when using data augmentation in Bayesian neural networks, as it changes the shape of the
posterior.

  \section{Related work}
In earlier sections of the paper, we already briefly discussed and reviewed relevant works on mode connectivity,
symmetries in the loss landscape and connection with gradient-based optimization methods. Here we discuss some relevant works
on the connection to Bayesian deep learning. The work of \citet{Garipov2018} sparked several contributions on exploiting
mode connectivity for ensembling models, which is akin to Bayesian model averaging. For example, in \citep{IzmailovMKGVW19}
the authors propose to ensemble models using curve subspaces to construct low-dimensional subspaces of parameter space.
These curve subspaces are, among others, the non-linear paths connecting low-loss modes (and consequently high-posterior
density) in weight space. In \citep{Fort2020}, the authors attempt to explain the effectiveness of deep ensembles \citep{Lakshminarayanan2017},
concluding that it is partially due to the diversity of the \gls{SGD} solutions in parameter space induced by random
initialization. More recently, in \citep{Benton2021a} the authors reason about mode connecting volumes, which are multi-dimensional
manifolds of low loss that connect many independently trained models.
These mode connecting volumes form the basis for an efficient method for building simplicial complexes for ensembling.
Here, we want to highlight that these works have not taken into account the permutation symmetries. Finally, in a concurrent
submission, \citep{Wiese2023} proposes an algorithm to remove symmetries in \gls{MCMC} chains for tanh networks.

\section{Conclusions}

By studying the effect of permutation symmetries, which are ubiquitous in neural networks, it is possible to analyze the
fundamental geometric properties of loss landscapes like (linear) mode connectivity and loss barriers. While previously
this was done on loss-optimized networks \citep{Ainsworth2023,Entezari2022,Frankle2020a}, in this work we have extended
the analysis to Bayesian neural networks. We have studied the linear connectivity properties of approximate Bayesian solutions
and we have proposed a matching algorithm (\cref{alg:align_vi}) to search for linearly connected solutions, by aligning
the distributions of two independent \gls{VI} solutions with respect to permutation matrices. We have empirically
validated our framework on a variety of experiments, showing that we can find zero barrier linearly-connected solutions for
\glspl{BNN} trained with \gls{VI}, on shallow models as well as on deep convolutional networks. This brings evidence for
\cref{claim:lmc} regarding the linear connectivity of approximate Bayesian solutions. Furthermore, we have studied the effect
of various design hyper-parameters, like width, prior and temperature, and observed complex patterns of behavior, which would
require additional research. In particular, the experiments raise questions regarding the relation between linear mode connectivity
and the generalization of \glspl{BNN}, as well as the role of width with respect to limiting behaviors of non-parametric
models like \glspl{GP} and \glspl{DGP}.

  \clearpage
  \begin{ack}
    SR wants to thank David Bertrand and the AIAO team at Stellantis for their work on the software and hardware infrastructure
    used for this work.
  \end{ack}

  \renewcommand{\bibname}{References}
  \bibliographystyle{abbrvnat_nourl}

  \appendix
  \renewcommand{\thefigure}{A\arabic{figure}}
  \cleardoublepage
  \section{Future work and open questions}

Besides the specific open questions discussed in the conclusions, we anticipate some possible future work. The framework we proposed to
analyze the loss barriers in \glspl{BNN} is general and can be applied to approximations other than \gls{VI}.
Future work could undertake to extend this analysis to the Laplace approximation \citep{MacKay1998}. However, this raises
methodological challenges since \cref{alg:align_vi} is not directly applicable, due to the use of a dense covariance
matrix (the inverse of the Hessian). Additionally, it would be worth extending these results to sample-based inference, like
\gls{SGHMC} \citep{Cheni2014}, or particle-based inference, like Stein variational inference \citep{Liu2016}. This would
call for a careful analysis, starting from the solution of \cref{eq:wasserstein_geodesics}, which would require approximations
\citep{Cuturi2013}.

\section{Additional details on the alignment method}
In the main paper, to align the distributions with respect to permutation matrices we argue to use the Wasserstein distance
rather than the \gls{KL} divergence. Indeed, by considering the \gls{KL} divergence $\KL{P_\#q_1}{q_0}$ between Gaussians
we have
\begin{align}
  \KL{P_\#q_1}{q_0} & = \logdet\diag(\mbs_{0}) - \logdet\diag(\mbP\mbs_{1}) + \Tr(\diag(\mbP\mbs_{1}\mbs_{0}^{\inv})) + \\
                    & (\mbm_{0}- \mbP\mbm_{1})^{\top}\diag(\mbs_{0}^{\inv})(\mbm_{0}- \mbP\mbm_{1})
\end{align}
It's easy to verify that the first three terms do not depend on $\mbP$, leading to just a distance between means and
disregarding any covariance information. In the figure below, we visualize the difference between doing \gls{LAP} with the
\gls{KL} cost and \gls{LAP} with the Wasserstein cost.
\begin{figure}[H]
  \centering
  \includegraphics[width=.9\textwidth]{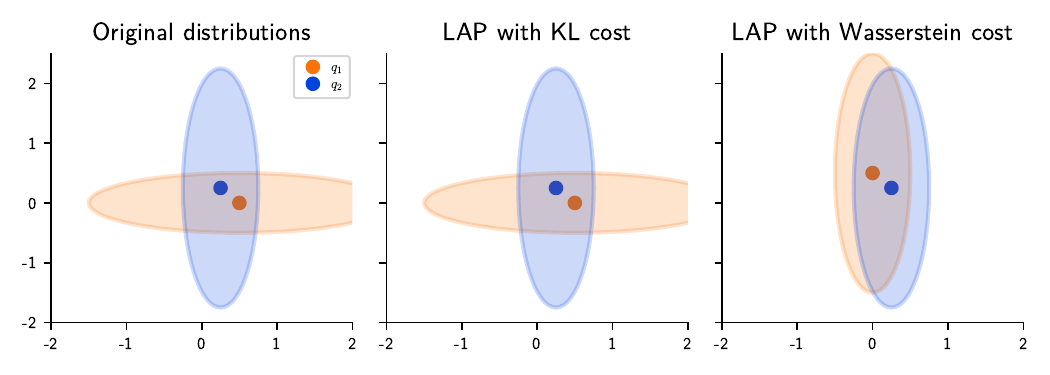}
  \caption{\textbf{Alignment using different objectives.} Given two distributions symmetrical w.r.t. the $y=x$ plane, using
  the \gls{KL} cost \gls{LAP} results in the identity permutation (which fails to recover the symmetry), while the
  Wasserstein cost better aligns the two distributions}
\end{figure}

\section{Experimental setup}
If not otherwise stated, we start by training two replicas of \glspl{BNN} with variational inference and we compute the marginalized
barrier as $\cB(q_{0}, q_{1}) = \max_{\tau}\cL(q_{\tau}) - \left( (1-\tau)\cL(q_{0}) + \tau\cL(q_{1}) \right)$ where $\cL
(\cdot)$ is the predictive likelihood and $\tau\in[0, 1]$, from which we take 25 evenly distributed points. In particular,
we seek to understand what happens to the \gls{VI} solutions with and without alignment applied to one of the two distribution
($q_{1}$ in our case).
We experiment with \glspl{MLP} with three layers and ResNet20 \citep{He2016} with various widths on MNIST \citep{mnist},
Fashion-MNIST \citep{fmnist} and CIFAR10 \citep{cifar}. All models are trained without data augmentation, because the
random augmentations introduce stochasticity which lacks a proper Bayesian interpretation in the formulation of the
likelihood function \citep{Wenzel2020}. Finally, we set the prior to be Gaussian $\cN(\mbzero, \alpha^{2}\mbI)$, with
the flexibility of choosing the variance. All \gls{VI} models are trained using the classic \gls{SGD} optimizer with momentum
\citep{Robbins1951,Rumelhart1986a} using the reparameterization trick \citep{Kingma14} with one sample during training and
128 samples during testing We use the categorical distribution and the Gaussian distribution as classification and regression
likelihood, respectively. \cref{tab:architecture_lenet,tab:architecture_resnet} show details on the \glspl{MLP} and
\glspl{CNN} base architectures used in our experimental campaign, while \cref{tab:hyperparams} reports the hyper-parameters
used in the experiments.
Note that differently from \citet{Entezari2022} and \citet{Ainsworth2023}, we don't use data augmentation. A possible protocol
for handling data augmentation in \glspl{BNN} is presented by \citet{Osawa2019} and involves carefully tuning the likelihood
temperature to correctly counting the number of data points.

\begin{table}[t]
	\footnotesize
\centering
\caption{Hyperparameters used for the experiments}
\label{tab:hyperparams}
\begin{tabular}{l|lll}
\toprule
\textbf{Dataset} & \multicolumn{2}{l}{CIFAR10} &     MNIST \\
\textbf{Model} &  ResNet20 &       MLP &       MLP \\
\midrule
Data Aug.     &     False &     False &     False \\
Batch size    &       500 &       500 &       500 \\
Temperature   &       1.0 &       1.0 &       1.0 \\
Test samples  &       128 &       128 &       128 \\
Train samples &         1 &         1 &         1 \\
VI std. init  &      0.01 &      0.01 &      0.01 \\
Base features &        16 &       512 &       512 \\
Prior var     &      0.01 &    0.0025 &      0.01 \\
Learning rate &  0.000001 &  0.000001 &  0.000001 \\
Train epochs  &      1000 &      1000 &      1000 \\
\bottomrule
\end{tabular}
\end{table}

\begin{table}[H]
  \centering
  \footnotesize
  \begin{minipage}[b]{0.45\textwidth}
    \centering
    \captionof{table}{MLP \label{tab:architecture_lenet}}
    \begin{tabular}{cc}
      \toprule \textbf{Layer} & \textbf{Dimensions} \\
      \hline
      \hline
      Linear-ReLU             & $512 \times \Din$   \\
      Linear-ReLU             & $512 \times 512$    \\
      Linear-ReLU             & $512 \times 512$    \\
      Linear-Softmax          & $\Dout \times 512$  \\
      \bottomrule
    \end{tabular}
  \end{minipage}
  \begin{minipage}[b]{0.45\textwidth}
    \centering
    \captionof{table}{ResNet20 \label{tab:architecture_resnet}}
    \begin{tabular}{cc}
      \toprule \textbf{Layer} & \textbf{Dimensions}                                                                  \\
      \hline
      \hline
      Conv2D                  & $16 \times 3 \times 3 \times \Din$                                                   \\
      \hline
      Residual Block          & $\left[ \begin{matrix}3 \times 3, 16 \\ 3 \times 3, 16\end{matrix} \right] \times 3$ \\
      \hline
      Residual Block          & $\left[ \begin{matrix}3 \times 3, 32 \\ 3 \times 3, 32\end{matrix} \right] \times 3$ \\
      \hline
      Residual Block          & $\left[ \begin{matrix}3 \times 3, 64 \\ 3 \times 3, 64\end{matrix} \right] \times 3$ \\
      \hline
      AvgPool                 & $8 \times 8$                                                                         \\
      Linear-Softmax          & $\Dout \times 64$                                                                    \\
      \bottomrule
    \end{tabular}
  \end{minipage}
\end{table}

\subsection{Computing platform}

The experiments have been performed using JAX \citep{JAX2018} and run on two AWS p4d.24xlarge instances with 8 NVIDIA A100
GPUs. Experiments were conducted using in the eu-west-1 region, which has a carbon efficiency of 0.62 kgCO$_{2}$eq/kWh.
A cumulative of 6500 hours of computation was performed on GPUs and it includes interactive sessions as well as small
experiments with very low GPU usage, providing a pessimistic estimation of the true utilization. Total emissions are
estimated to be 1007.5 kgCO$_{2}$eq of which 100 percents were directly offset by AWS.

\section{Additional results}

\begin{figure}[]
  \centering
  \includegraphics[width=.35\textwidth]{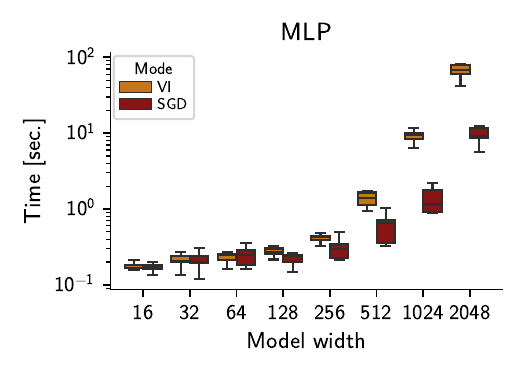}
  \includegraphics[width=.35\textwidth]{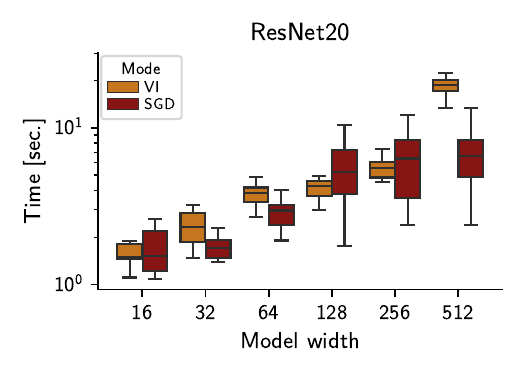}
  \caption{\textbf{Timings.} Profile of the algorithms to align the \gls{VI} solutions and to match weights from \gls{SGD}
  solutions.}
  \label{fig:timings}
\end{figure}

We present timings obtained by profiling the time needed to solve the \gls{SOLAP} with the Wasserstein cost, as well as the
time for the deterministic case \citep{Ainsworth2023}. In \cref{fig:timings} we show the results for MLP and ResNet20 architectures,
varying the model width. It is evident that, in the majority of cases, the algorithm completes within a minute. Moreover,
as anticipated, in case of \gls{VI} solving our distribution alignment problem for wide neural networks is more computationally
demanding compared to merely matching weights from \gls{SGD} solutions.

\begin{figure}[htbp]
  \centering
  \includegraphics[width=.35\textwidth]{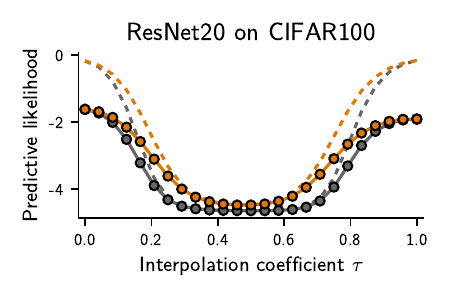}
  \includegraphics[width=.3\textwidth]{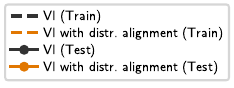}
  \caption{\textbf{Alignment failure.} The method proposed fails to recover zero-barrier solutions for CIFAR100.}
  \label{fig:cifar100}
\end{figure}

Finally, we also test our setup on the CIFAR100 dataset \citep{cifar}. Surprisingly, we were not able to replicate the same
level of performance as in the other cases. In \cref{fig:cifar100}, we see that, despite converging well, we fall short to
find zero-barrier solutions. Similarly to the comments of \citet{Ainsworth2023}, we also stress that the failure to
align distributions does not rule out the existence of a proper permutation map that the algorithm couldn't find. Nonetheless,
this raises a number of questions: the Bayesian posterior is the product of two ingredients, the prior and the
likelihood, conditioned to observing a dataset.

\begin{minipage}{.55\textwidth}
  Finally, as an additional check, we analyze the log-posterior by projecting the density into two dimensional slices,
  following the setup in \citep{Izmailov21,Garipov2018}. We study the two dimensional subspace of the parameter space
  supported by the hyperplane $H$ of the form
  \begin{align*}
    H = \{\mbtheta \in\bbR^{d}\g \mbtheta = a\mbtheta_{a}+ b\mbtheta_{b}+ (1 - a - b)\mbtheta_{c}\}\,,
  \end{align*}
  where $a,b\in\bbR$ and $\mbtheta_{a}$, $\mbtheta_{b}$ and $\mbtheta_{c}$ are the means of $q_{0}$, $q_{1}$ and
  $P_{\#}q_{1}$. With this configuration, all three solutions lie on this hyper-plane. In \cref{fig:log_posterior_2d},
  we present the visualization of ResNet20 trained on CIFAR10. We see that the distributions $q_{0}$ and $P_{\#}q_{1}$ are
  connected by higher density regions than the ones between $q_{0}$ and $q_{1}$. Also, as expected the symmetries arise from
  the form of the likelihood, and the prior has a comparable strength with respect to the three posteriors. Later, we
  also study in more details the effect of the prior's variance in finding low-barrier solutions.
\end{minipage}\hfill%
\begin{minipage}{.4\textwidth}
  \begin{minipage}[t]{0.5\textwidth}
    \vspace{0pt}
    \centering
    \adjincludegraphics[width=\textwidth,trim={0 0 {.66\width} 0},clip]{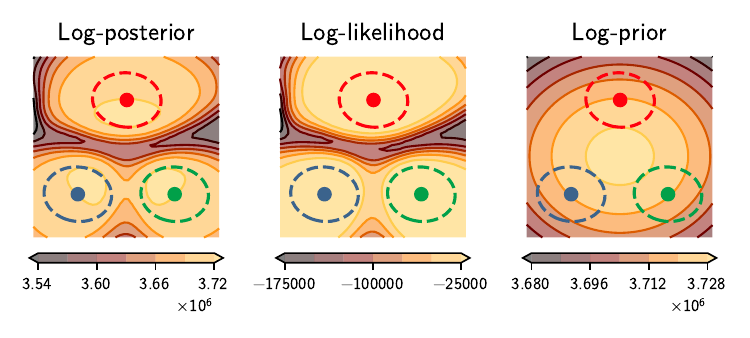}%
  \end{minipage}%
  \begin{minipage}[t]{0.35\textwidth}
    \vspace{5ex}
    \includegraphics[width=\textwidth]{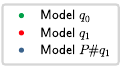}
  \end{minipage}\\[-3.5ex] \adjincludegraphics[width=\textwidth,trim={{.33\width} 0 0 0},clip]{figures/posterior_viz_resnet20.pdf}\\%
  \vspace{-5ex}
  \captionof{figure}{\textbf{Posterior density visualization.}
  All three solutions are local approximation of posterior, but $q_{0}$ and $P_{\#}q_{1}$ are connected by lower density regions.}
  \label{fig:log_posterior_2d_appendix}
\end{minipage}

  \section{A primer on variational inference for Bayesian neural networks}

\Gls{VI} is a classic tool to tackle intractable Bayesian inference \citep{Jordan1999,Blei2017}. \gls{VI} casts the inference
problem into an optimization-based procedure to compute a tractable approximation of the true posterior. Assume a
generic parametric model $f$ parameterized by some unknown parameters $\mbtheta$ (i.e. $f(\cdot,\mbtheta)$) and a collection
of data $\mby\in\bbR^{N}$ corresponding to some input points $\mbX=\{\mbx_{i} \g \mbx_{i}\in\bbR^{\Din} \}_{i=1,\dots, N}$.
In our setting, we have a probabilistic model $p(\mby\g f(\mbX; \mbtheta))$ with parameters $\mbtheta$, a prior
distributions on them $p(\mbtheta)$ and a set of observations $\{\mbX,\mby\}$.
In a nutshell, the general recipe of \gls{VI} consists of (i) introducing a set $\cQ$ of distributions; (ii) defining a
tractable objective that ``measure'' the distance between any arbitrary distribution $q(\mbtheta)\in\cQ$ and the true
posterior $p(\mbtheta\g\mby)$; and finally (iii) providing a programmatic way to find the distribution $\widetilde{q}(\mbtheta
)$ that minimizes such distance. In practice, $q(\mbtheta)$ has some free parameters $\mbnu$ (also known as \textit{variational
parameters}), which are optimized such that the approximating distribution $q(\mbtheta;\mbnu)$ is as closer as possible to
the true posterior $p(\mbtheta\g\mby)$.
We can derive the variational objective starting from the definition of the \gls{KL},
\begin{align}
  \label{eq:derivation_bound}{\KL{q(\mbtheta;\mbnu)}{p(\mbtheta\g\mby)}} & = \bbE_{q(\mbtheta;\mbnu)}\left[\log q(\mbtheta;\mbnu) - \log p(\mbtheta\g\mby) \right] =                                        \\
                                                                         & = \E_{q(\mbtheta;\mbnu)}\left[\log q(\mbtheta;\mbnu) - \log p(\mby\g\mbtheta) - \log p(\mbtheta)\right] +{\log p(\mby)}\nonumber %
\end{align}
Rearranging we have that
\begin{align}
  \label{eq:derivation_bound2}{\log p(\mby) - \KL{q(\mbtheta;\mbnu)}{p(\mbtheta\g\mby)}}= \E_{q(\mbtheta;\mbnu)}\left[\log q(\mbtheta;\mbnu) - \log p(\mby\g\mbtheta) - \log p(\mbtheta)\right]
\end{align}
The r.h.s. of the equation defines our variational objective, also known as \gls{ELBO}, that can be arranged as follows,
\begin{align}
  \label{eq:elbo_full}\Lelbo(\mbnu) = \underbrace{\E_{q(\mbtheta;\mbnu)} \log p(\mby\g\mbtheta)}_{\text{Model fitting term}}- \underbrace{\KL{q(\mbtheta;\mbnu)}{p(\mbtheta)}}_{\text{Regularization term}}\,.
\end{align}
This formulation highlights the property of this objective, which is made of two components: the first one is the expected
log-likelihood under the approximate posterior $q$ and measures how the model fits the data. The second term, on the other
hand, has the regularization effect of penalizing posteriors that are far from the prior as measured by the \gls{KL}.
Before diving into the challenges of optimization of the \gls{ELBO}, we shall spend a brief moment discussing the form
of the approximating distribution $q$. One of the simplest and easier choice is the mean field approximation \citep{Hinton1993},
where each variable $\theta_{i}$ is taken to be independent with respect to the remaining $\mbtheta_{-i}$. Effectively, this
imposes a factorization of the posterior,
\begin{equation}
  q(\mbtheta;\mbnu) = \prod_{i=1}^{K} q(\theta_{i};\mbnu_{i})
\end{equation}
where $\mbnu_{i}$ is the set of variational parameters for the parameter $\theta_{i}$. On top of this approximation,
$q(\theta_{i})$ is often chosen to be Gaussian,
\begin{equation}
  q(\theta_{i}) = \cN(\mu_{i}, \sigma^{2}_{i})
\end{equation}
Now, the collection of all means and variances $\{\mu_{i},\sigma_{i}^{2}\}_{i=1}^{K}$ defines the set of variational
parameters to optimize.

For \glspl{BNN} the analytic evaluation of the \gls{ELBO} (and its gradients) is always untractable due the non-linear
nature of the expectation of the log-likelihood under the variational distribution. Nonetheless, this can be easily
estimated via Monte Carlo integration \citep{Metropolis1949}, by sampling $\nmc$ times from $q_{\mbnu}$,
\begin{align}
  \label{eq:expected-loglikelihood}\E_{q(\mbtheta;\mbnu)}\log p(\mby\g\mbtheta) \approx \frac{1}{\nmc}\sum_{j=1}^{\nmc} \log p(\mby\g\widetilde\mbtheta_{j})\,,\quad\text{with}\quad\widetilde\mbtheta_{j}\sim q(\mbtheta;\mbnu)
\end{align}
In practice, this is as simple as re-sampling the weights and the biases for all the layers $\nmc$ times and computing
the output for each new sample.

We now have a tractable objective that needs to be optimized with respect to the variational parameters $\mbnu$. Very
often the \gls{KL} term is known, making its differentation trivial. On the other hand the expectation of the likelihood
is not available, making the computation of its gradients more challenging.
This problem can be solved using the so-called \textit{reparameterization trick} \citep{Salimans2013, Kingma14}. The
reparameterization trick aims at constructing $\mbtheta$ as an invertible function $\cT$ of the variational parameters
$\mbnu$ and of another random variable $\mbvarepsilon$, so that $\mbtheta = \mathcal{T}(\mbvarepsilon; \mbnu)$.
Generally, a $\mathcal{T}$ that suits this constraint might not exists; \citet{Ruiz2016} discuss how to build ``weakly''
dependent transformation $\mathcal{T}$ for distributions like Gamma, Beta and Log-normal. For discrete distributions,
instead, one could use a continuous relaxation, like the Concrete \citep{Maddison2017}. $\mbvarepsilon$ is chosen such
that its marginal $p(\mbvarepsilon)$ does not depend on the variational parameters. With this parameterization,
$\mathcal{T}$ separates the deterministic components of $q$ from the stochastic ones, making the computation of its gradient
straightforward.
For a Gaussian distribution with mean $\mu$ and variance $\sigma^{2}$, $\mathcal{T}$ corresponds to as simple scale-location
transformation of an isotropic Gaussian noise,
\begin{align}
  \label{eq:reparameterization-trick-gaussian}\theta \sim \cN(\mu, \sigma^{2}) \iff \theta = \mu + \sigma\varepsilon \quad \text{with}\quad \varepsilon \sim \cN(0, 1)\,.
\end{align}
This simple transformation ensures that $p(\varepsilon) = \cN(0, 1)$ does not depends on the variational parameters
$\mbnu = \{\mu, \sigma^{2}\}$. The gradients of the \gls{ELBO} can be therefore computed as
\begin{align}
  \label{eq:gradients-elbo}\gradient_{\mbnu}\Lelbo = \E_{p(\mbvarepsilon)}\left[\gradient_{\mbtheta} \log p(\mby\g\mbtheta)\g_{\mbtheta = \mathcal{T}(\mbvarepsilon; \mbnu)}\gradient_{{\mbnu}}\mathcal{T}(\mbvarepsilon; \mbnu) \right] - \gradient_{\mbnu} \KL{q(\mbtheta; \mbnu)}{p(\mbtheta)}\,.
\end{align}
The gradient $\gradient_{\mbtheta} \log p(\mby\g\mbtheta)$ depends on the model and it can be derived with automatic
differentation tools \citep{Tensorflow2015,Pytorch2019}, while $\gradient_{{\mbnu}}\mathcal{T}(\mbvarepsilon; \mbnu)$
doesn't have any stochastic components and therefore can be known deterministically. Note that the reparameterization trick
can be also used when the \gls{KL} is not analitically available. In that case, we would end up with,
\begin{align}
  \label{eq:gradients-elbo-full-reparam}\gradient_{\mbnu}\Lelbo = \E_{p(\mbvarepsilon)}\left[\gradient_{\mbtheta} \log p(\mby\g\mbtheta) + \log q(\mbtheta;\mbnu) - \log p(\mbtheta)\right]_{\mbtheta = \mathcal{T}(\mbvarepsilon; \mbnu)}\gradient_{{\mbnu}}\mathcal{T}(\mbvarepsilon; \mbnu)
\end{align}
\citet{Roeder2017} argue that when we believe that $q(\mbtheta;\mbnu)\approx p(\mby\g\mbtheta)$, \cref{eq:gradients-elbo-full-reparam}
should be prefered over \cref{eq:gradients-elbo} even if computing analitically the \gls{KL} is possible. Note that this
case is very unlikely for \gls{BNN} posteriors, and that the additional randomness introduced by the Monte Carlo estimation
of the \gls{KL} could be harmful.

In case of large datasets and complex models, the formulation summarized in \cref{eq:gradients-elbo} can be
computationally challenging, due to the evaluation of the likelihood and its gradients $\nmc$ times. Assuming
factorization of the likelihood,
\begin{equation}
  p(\mby\g\mbtheta) = p(\mby\g f(\mbX;\mbtheta)) = \prod_{i=1}^{N} p(y_{i}\g f(\mbx_{i};\mbtheta))
\end{equation}
this quantity can be approximated using mini-batching \citep{Graves2011,Hoffman2013}. Recalling $\mby$ as the set of labels
of our dataset with $N$ examples, by taking $\cB \subset \mby$ as a random subset of $\mby$, the likelihood term can be estimated
in an unbiased way as
\begin{align}
  \label{eq:likelihood-with-minibatch}\log p_{\mbtheta}(\mby\g\mbtheta) \approx \frac{N}{M}\sum_{y_i\sim\cB}\log p(y_{i}\g\mbtheta)\,. %
\end{align}
where $M$ is the number of points in the minibatch. At the cost of increase ``randomness'', we can use \cref{eq:gradients-elbo}
to compute the gradients of the \gls{ELBO} with the minibatch formulation in \cref{eq:likelihood-with-minibatch}. Stochastic
optimization, e.g. any version of \gls{SGD}, will converge to a local optimum provided with a decreasing learning rate and
sufficient gradient updates \citep{Robbins1951}.

\end{document}